\documentclass[10pt,journal,compsoc]{IEEEtran}
\ifCLASSOPTIONcompsoc
  \usepackage[nocompress]{cite}
\else
  \usepackage{cite}
\fi
\ifCLASSINFOpdf
\else
\fi
\usepackage{algorithmic}
\usepackage{fixltx2e}
\usepackage{bm}
\usepackage{graphicx}
\usepackage{mathtools,amsmath}
\usepackage{amsfonts}
\usepackage{color,soul}
\usepackage{hyperref}

\usepackage{booktabs}
\newcommand{\ra}[1]{\renewcommand{\arraystretch}{#1}}
\usepackage{tabu}
\usepackage{multirow}

\newcommand{\norm}[1]{\left\lVert#1\right\rVert}

\usepackage{xspace}

\makeatletter
\DeclareRobustCommand\onedot{\futurelet\@let@token\@onedot}
\def\@onedot{\ifx\@let@token.\else.\null\fi\xspace}

\def\eg{\emph{e.g}\onedot} 
\def\ie{\emph{i.e}\onedot} 
 
\def\etc{\emph{etc}\onedot} 
 
\def\etal{\emph{et al}\onedot}

\makeatother

\newcommand{\mysubsec}[1]{\vspace{1mm}\noindent{\bf #1}}

\begin{document}
%
\title{Generative Adversarial Networks for Image and Video Synthesis: Algorithms and Applications}
%
%
%
%

\author{
Ming-Yu Liu*,
Xun Huang*,
Jiahui Yu*,
Ting-Chun Wang*,
Arun Mallya*
\IEEEcompsocitemizethanks{
\IEEEcompsocthanksitem * Equal contribution. Ming-Yu Liu, Xun Huang, Ting-Chun Wang, and Arun Mallya are with NVIDIA. Jiahui Yu is with Google.\protect
}
}

\IEEEtitleabstractindextext{%
\begin{abstract}
The generative adversarial network (GAN) framework has emerged as a powerful tool for various image and video synthesis tasks, allowing the synthesis of visual content in an unconditional or input-conditional manner. It has enabled the generation of high-resolution photorealistic images and videos, a task that was challenging or impossible with prior methods. It has also led to the creation of many new applications in content creation. In this paper, we provide an overview of GANs  with a special focus on algorithms and applications for visual synthesis. We cover several important techniques to stabilize GAN training, which has a reputation for being notoriously difficult. We also discuss its applications to image translation, image processing, video synthesis, and neural rendering.
\end{abstract}

}

\maketitle

\IEEEdisplaynontitleabstractindextext

%
\IEEEpeerreviewmaketitle

\IEEEraisesectionheading{\section{Introduction}\label{sec:introduction}}

\IEEEPARstart{T}{he} generative adversarial network (GAN) framework is a deep learning architecture~\cite{lecun2015deep,goodfellow2016deep} introduced by Goodfellow~\etal~\cite{goodfellow2014generative}. It consists of two interacting neural networks---a generator network $G$ and a discriminator network $D$, which are trained jointly by playing a zero-sum game where the objective of the generator is to synthesize fake data that resembles real data, and the objective of the discriminator is to distinguish between real and fake data. When the training is successful, the generator is an approximator of the underlying data generation mechanism in the sense that the distribution of the fake data converges to the real one. Due to the distribution matching capability, GANs have become a popular tool for various data synthesis and manipulation problems, especially in the visual domain.

GAN's rise also marks another major success of deep learning in replacing hand-designed components with machine-learned components in modern computer vision pipelines. As deep learning has directed the community to abandon hand-designed features, such as the histogram of oriented gradients (HOG)~\cite{dalal2005histograms}, for deep features computed by deep neural networks, the objective function used to train the networks remains largely hand-designed. While this is not a major issue for a classification task since effective and descriptive objective functions such as the cross-entropy loss exist, this is a serious hurdle for a generation task. After all, how can one hand-design a function to guide a generator to produce a better cat image? How can we even mathematically describe ``\emph{felineness}'' in an image?

\begin{figure}[t!]
    \centering
    \includegraphics[width=0.45\textwidth, trim=0 0 0 0, clip]{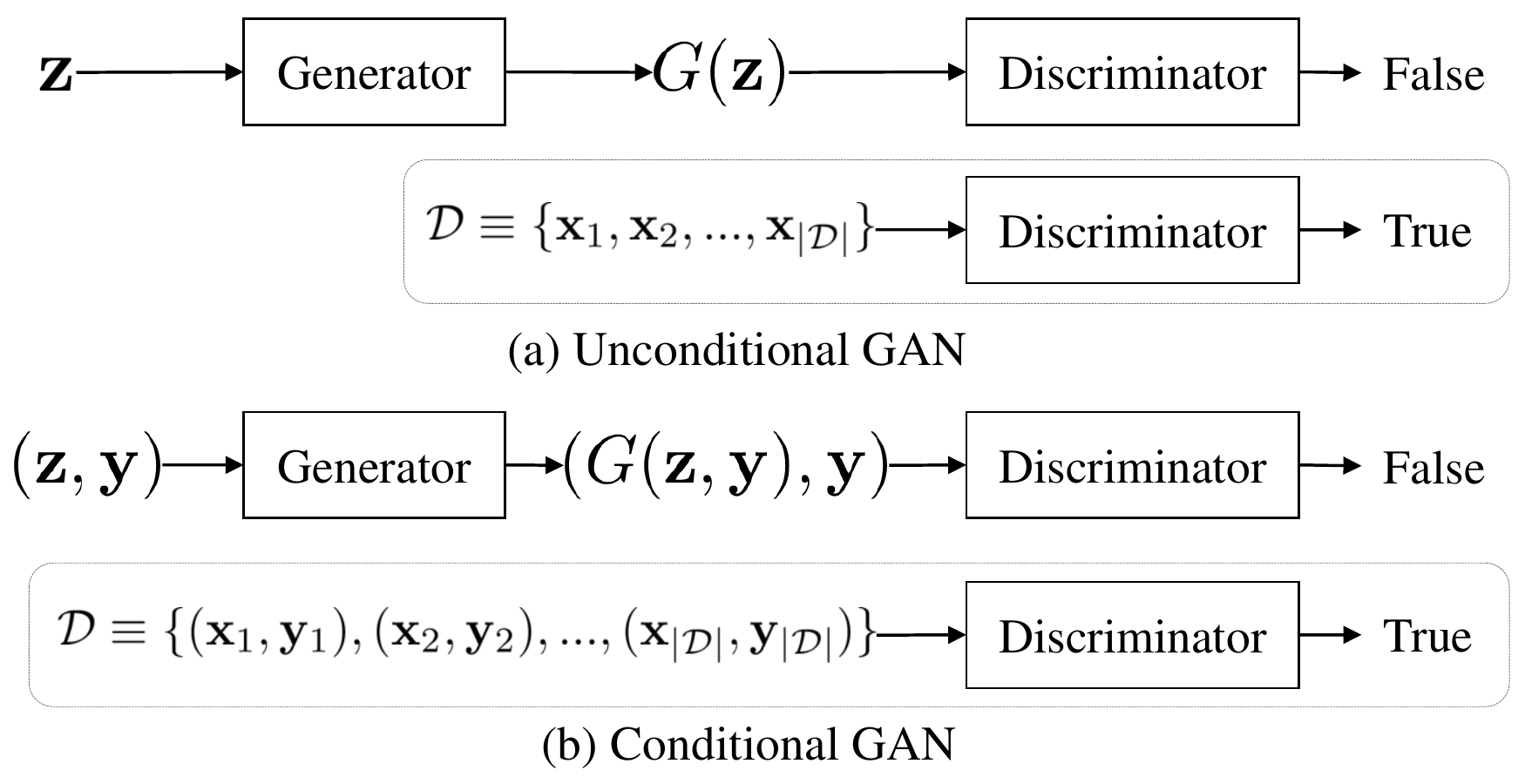}
    \caption{{\bf Unconditional vs. Conditional GANs.} (a) In unconditional GANs, the generator converts a noise input $\bm{z}$ to a fake image $G(\bm{z})$ where $\bm{z}\sim \mathcal{Z}$ and $\mathcal{Z}$ is usually a Gaussian random variable. The discriminator tells apart real images $\bm{x}$ from the training dataset $\mathcal{D}$ and fake images from $G$. (b) In conditional GANs, the generator takes an additional input $\bm{y}$ as the control signal, which could be another image (image-to-image translation), text (text-to-image synthesis), or a categorical label (label-to-image synthesis). The discriminator tells apart real from fake by leveraging the information in $\bm{y}$. In both settings, the combination of the discriminator and real training data defines an objective function for image synthesis. This data-driven objective function definition is a powerful tool for many computer vision problems.}
    \label{fig:gan_vs_cgan}
\end{figure}

GANs address the issue through deriving a functional form of the objective using training data. As the discriminator is trained to tell whether an input image is a cat image from the training dataset or one synthesized by the generator, it defines an objective function that can guide the generator in improving its generation based on its current network weights. The generator can keep improving as long as the discriminator can differentiate real and fake cat images. The only way that a generator can beat the discriminator is to produce images similar to the real images used for training. Since all the training images contain cats, the generator output must contain cats to win the game. Moreover, when we replace the cat images with dog images, we can use the same method to train a dog image generator. The objective function for the generator is defined by the training dataset and the discriminator architecture. It is thus a very flexible framework to define the objective function for a generation task as illustrated in Figure~\ref{fig:gan_vs_cgan}.

\begin{figure}[!t]
    \centering
    \includegraphics[width=\columnwidth, trim=0.1cm 0 0 0, clip]{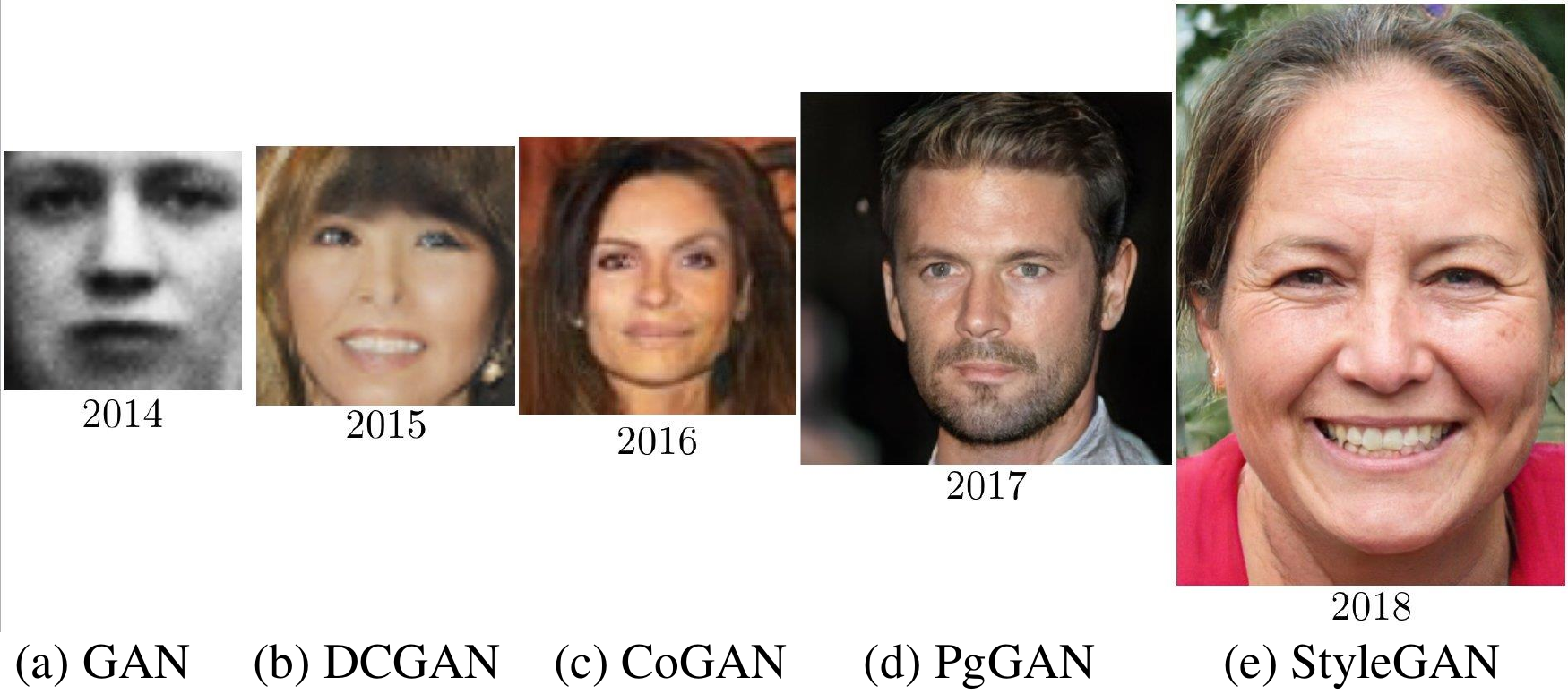}
    \caption{{\bf GAN progress on face synthesis.} The figure shows the progress of GANs on face synthesis over the years. From left to right, we have face synthesis results by (a) the original GAN~\cite{goodfellow2014generative},  DCGAN~\cite{radford2015unsupervised},  CoGAN~\cite{liu2016coupled},  PgGAN~\cite{karras2017progressive}, and  StyleGAN~\cite{karras2018style}. This image was originally created and shared by Ian Goodfellow on \href{https://twitter.com/goodfellow_ian/status/1084973596236144640}{\ul{Twitter}}.}
    \label{fig:progress}
\end{figure}

However, despite its excellent modeling power, GANs are notoriously difficult to train because it involves chasing a moving target. Not only do we need to make sure the generator can reach the target, but also that the target can reach a desirable level of goodness. Recall that the goal of the discriminator is to differentiate real and fake data. As the generator changes, the fake data distribution changes as well. This poses a new classification problem to the discriminator, distinguishing the same real but a new kind of fake data distribution, one that is presumably more similar to the real data distribution. As the discriminator is updated according to the new classification problem, it induces a new objective for the generator. Without careful control of the dynamics, a learning algorithm tends to experience failures in GAN training. Often, the discriminator becomes too strong and provides strong gradients that push the generator to a numerically unstable region. This is a well-recognized issue. Fortunately, over the years, various approaches, including better training algorithms, network architectures, and regularization techniques, have been proposed to stabilize GAN training. We will review several representative approaches in Section~\ref{sec:learning}. In Figure~\ref{fig:progress}, we illustrate the progress of GANs over the past few years.


In the original GAN formulation~\cite{goodfellow2014generative}, the generator is formulated as a mapping function that converts a simple, unconditional distribution, such as a uniform distribution or a Gaussian distribution, to a complex data distribution, such as a natural image distribution. We now generally refer to this formulation as the unconditional GAN framework. While the unconditional framework has several important applications on its own, the lack of controllability in the generation outputs makes it unfit to many applications. This has motivated the development of the conditional GAN framework. In the conditional framework, the generator additionally takes a control signal as input. The signal can take in many different forms, including category labels, texts, images, layouts, sounds, and even graphs. The goal of the generator is to produce outputs corresponding to the signal. In Figure~\ref{fig:gan_vs_cgan}, we compare these two frameworks. This conditional GAN framework has led to many exciting applications. We will cover several representative ones through Sections~\ref{sec:supervised}~to~\ref{sec:neural_rendering}.

GANs have led to the creation of many exciting new applications. For example, it has been the core building block to semantic image synthesis algorithms that concern converting human-editable semantic representations, such as segmentation masks or sketches, to photorealistic images. GANs have also led to the development of many image-to-image translation methods, which aim to translate an image in one domain to a corresponding image in a different domain. These methods find a wide range of applicability, ranging from image editing to domain adaptation. We will review some algorithms in this space in Section~\ref{sec:supervised}.

We can now find GAN's footprint in many visual processing systems. For example, for image restoration, super-resolution, and inpainting, where the goal is to transform an input image distribution to a target image distribution, GANs have been shown to generate results with much better visual quality than those produced with traditional methods. 
We will provide an overview of GAN methods in these image processing tasks in Section~\ref{sec:image_processing}.

Video synthesis is another exciting area that GANs have shown promising results. Many research works have utilized GANs to synthesize realistic human videos or transfer motions from one person to another for various entertainment applications, which we will review in Section~\ref{sec:video}. Finally, thanks to its great capability in generating photo-realistic images, GANs have played an important role in the development of neural rendering---using neural networks to boost the performance of the graphics rendering pipeline. We will cover GAN works in this space in Section~\ref{sec:neural_rendering}.

\section{Related Works}\label{sec:related}

\begin{figure*}[tbh!]
    \centering
    \includegraphics[width=\textwidth, trim=0 0 0 0, clip]{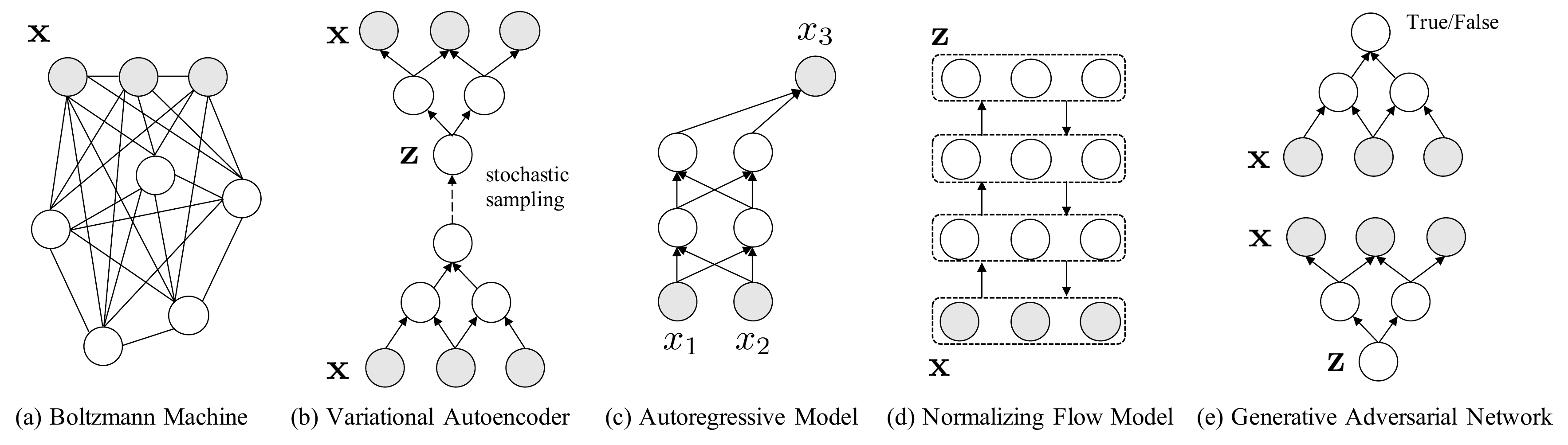}
    \caption{{\bf Structure comparison of different deep generative models.} Except the deep Boltzmann machine which is based on undirected graphs, the other models are all based on directed graphs, which enjoy a faster inference speed.}
    \label{fig:method_overview}
\end{figure*}

Several GAN review articles exist, including the introductory article by Goodfellow~\cite{goodfellow2016nips}. The articles by Creswell~\etal~\cite{creswell2018generative} and Pan~\etal~\cite{pan2019recent} summarize GAN methods prior to 2018. Wang~\etal~\cite{wang2019generative} provides a taxonomy of GANS. Our work differs from the prior works in that we provide a more contemporary summary of GAN works with a focus on image and video synthesis.

There are many different deep generative models or deep neural networks that model the generation process of some data. Besides GANs, other popular deep generative models include deep Boltzmann Machines, variational autoencoders, deep autoregressive models, and normalizing flow models. We compare these models in Figure~\ref{fig:method_overview} and briefly review them below.

\mysubsec{Deep Boltzmann Machines (DBMs).} DBMs~\cite{hinton2006reducing,salakhutdinov2009deep,fischer2012introduction,du2019implicit} are energy-based models~\cite{lecun2006tutorial}, which can be represented by undirected graphs. Let $\bm{x}$ denote the array of image pixels, often called visible nodes. Let $\bm{h}$ denote the hidden nodes. DBMs model the probability density function of data based on the Boltzmann (or Gibbs) distribution as
\begin{equation}
p(\bm{x}; \bm{\theta}) = \frac{1}{N(\bm{\theta})}\sum_{\bm{h}} \exp(-E(\bm{x}, \bm{h}; \bm{\theta})),
\end{equation}
where $E$ is an energy function modeling interactions of nodes in the graph, $N$ is the partition function, and $\bm{\theta}$ denotes the network parameters to be learned. Once a DBM is trained, a new image can be generated by applying Markov Chain Monte Carlo (MCMC) sampling, ascending from a random configuration to one with high probability. While extensively expressive, the reliance on MCMC sampling on both training and generation makes DBMs scale poorly compared to other deep generative models, since efficient MCMC sampling is itself a challenging problem, especially for large networks.

\mysubsec{Variational AutoEncoders (VAEs).} VAEs~\cite{kingma2019introduction,kingma2013auto,rezende2014stochastic} are directed probabilistic graphic models, inspired by the Helmholtz machine~\cite{dayan1995helmholtz}. They are also descendant of latent variable models, such as principal component analysis and autoencoders~\cite{bengio2013representation}, which concern representing high-dimensional data $\bm{x}$ using lower-dimensional latent variables $\bm{z}$. In terms of structure, a VAE employs an inference model $q(\bm{z}|\bm{x};\bm{\phi})$ and a generation model $p(\bm{x}|\bm{z};\bm{\theta})p(\bm{z})$ where $p(\bm{z})$ is usually a Gaussian distribution, which we can easily sample from, and $q(\bm{z}|\bm{x};\bm{\phi})$ approximates the posterior $p(\bm{z}|\bm{x};\bm{\theta})$. Both of the inference and generation models are implemented using feed-forward neural networks. VAE training is through maximizing the evidence lower bound (ELBO) of $\log p(\bm{x};\bm{\theta})$ and the non-differentiblity of the stochastic sampling is elegantly handled by the reparametrization trick~\cite{kingma2019introduction}.  One can also show that maximization the ELBO is equivalent to minimizing the Kullback–Leibler (KL) divergence 
\begin{equation}\label{eqn::vae_formulation}
KL\left(q(\bm{x})q(\bm{z}|\bm{x};\bm{\phi})|| p(\bm{z})p(\bm{x}|\bm{z};\bm{\theta} )\right),
\end{equation}
where $q(x)$ is the empirical distribution of the data~\cite{kingma2019introduction}. Once a VAE is trained, an image can be efficiently generated by first sampling $\bm{z}$ from the Gaussian prior $p(\bm{z})$ and then passing it through the feed-forward deep neural network $p(\bm{x}|\bm{z};\bm{\theta})$. VAEs are effective in learning useful latent representations~\cite{sonderby2016ladder}. However, they tend to generate blurry output images.

\mysubsec{Deep AutoRegressive Models (DARs).} DARs~\cite{van2016conditional,oord2016wavenet,salimans2017pixelcnn++,chen2018pixelsnail} are deep learning implementations of classical autoregressive models, which assume an ordering to the random variables to be modeled and generate the variables sequentially based on the ordering. This induces a factorization form to the data distribution given by solving
\begin{equation}
p(\bm{x};\bm{\theta})=\prod_{i} p\left(x_i|\bm{x}_{<i};\bm{\theta}\right),
\end{equation}
where $x_i$s are variables in $\bm{x}$, and $\bm{x}_{<i}$ are the union of the variables that are prior to $x_i$ based on the assumed ordering. DARs are conditional generative models where they generate a new portion of the signal based on what has been generated or observed so far. The learning is based on maximum likelihood learning
\begin{equation}
\max_{\bm{\theta}} \mathbb{E}_{\bm{x}\sim\mathcal{D}}\left[\log p(x_i|\bm{x}_{<i};\bm{\theta}) \right].
\end{equation}
DAR training is more stable compared to the other generative models. But, due to the recurrent nature, they are slow in inference. Also, while for audio or text, a natural ordering of the variables can be determined based on the time dimension, such an ordering does not exist for images. One hence has to enforce an order prior that is an unnatural fit to the image grid.

\mysubsec{Normalizing Flow Models (NFMs).} NFMs~\cite{rezende2015variational,dinh2014nice,dinh2016density,kingma2018glow} are based on the normalizing flow---a transformation of a simple probability distribution into a more complex distribution by a sequence of invertible and differentible mappings. Each mapping corresponds to a layer in a deep neural network. With a layer design that guarantees invertibility and differentibility for all possible weights, one can stack many such layers to construct a powerful mapping because composition of invertible and differentible functions are invertible and differentible. Let $F=f^{(1)}\circ f^{(2)}...\circ f^{(K)}$ be such a $K$-layer mapping that maps the simple probability distribution $Z$ to the data distribution $X$. The probability density of a sample $\bm{x}\sim X$ can be computed by transforming it back to the corresponding $\bm{z}$. Hence, we can apply maximum likelihood learning to train NFMs because the log-likelihood of the complex data distribution can be converted to the log-likelihood of the simple prior distribution subtracted by the Jacobians terms. This gives
\begin{equation}
\log p(\bm{x};\bm{\theta}) = \log p(\bm{z};\bm{\theta}) - \sum_{i=1}^K \log \left| \det \frac{d f^{(i)}}{d \bm{z}_{i-1}} \right|,
\end{equation}
where $\bm{z}_i=f^{(i)}(\bm{z}_{i-1})$. One key strength of NFMs is in supporting direct evaluation of probability density calculation. However, NFMs require an invertible mapping, which greatly limits the choices of applicable architectures.

\section{Learning}\label{sec:learning}

Let $\bm{\theta}$ and $\bm{\phi}$ be the learnable parameters in $G$ and $D$, respectively. GAN training is formulated as a minimax problem
\begin{equation}
    \min_{\bm{\phi}} \max_{\bm{\theta}} V(\bm{\theta}, \bm{\phi}),
\end{equation}
where $V$ is the utility function. 

GAN training is challenging. Famous failure cases include mode collapse and mode dropping. In mode collapse, the generator is trapped to a certain local minimum where it only captures a small portion of the distribution. In mode dropping, the generator does not faithfully model the target distribution and misses some portion of it. Other common failure cases include checkerboard and waterdrop artifacts. In this paper, we cover the basics of GAN training and some techniques invented to improve training stability.

\subsection{Learning Objective}

\begin{table}[!t]
    \ra{1.3}
    \centering
    \caption{{\bf Comparison of different GAN losses}, including saturated~\cite{goodfellow2014generative}, non-saturated~\cite{goodfellow2014generative}, Wasserstein~\cite{arjovsky2017wasserstein}, least square~\cite{mao2017least}, and hinge~\cite{lim2017geometric,zhang2019self}, in terms of the discriminator output layer type in~(\ref{eqn:dis_update}) and~(\ref{eqn:gen_update}). We maximize
    	$f_D$ and $f_G$ for training the discriminator. As shown in (\ref{eqn:dis_update}) and (\ref{eqn:gen_update}), we minimize $g_G$ for training the generator. Note that $\sigma(x)=\frac{1}{1+e^{-x}}$ is the sigmoid function.}\label{tbl:learning_objective}
	\begin{tabular}{@{}rccc}
		\toprule
		Loss & $f_D(x)$ & $f_G(x)$ & $g_G(x)$\\\midrule
		Saturated & 
		$\log \sigma(x)$& 
		$\log (1 - \sigma(x))$& 
		$\log (1 - \sigma(x))$\\
		Non-Saturated & 
		$\log \sigma(x)$& 
		$\log (1 - \sigma(x))$& 
		$-\log \sigma(x)$\\
		Wasserstein & 
		$x$ & 
		$-x$ & 
		$-x$\\			
		Least-Square & 
		$-(x-1)^2$&
		$-x^2$& 
		$(x-1)^2$\\					
		Hinge & 
		$\min(0, x-1)$& 
		$\min(0, -x-1)$& 
		$-x$\\\bottomrule
	\end{tabular}
\end{table}

The core idea in GAN training is to minimize the discrepancy between the true data distribution $p(\bm{x})$ and the fake data distribution $p(G(\bm{z};\bm{\theta}))$. As there are a variety of ways to measure the distance between two distributions, such as the Jensen-Shannon divergence, the Kullback$-$Leibler divergence, and the integral probability metric, there are also a variety of GAN losses, including the saturated GAN loss~\cite{goodfellow2014generative}, the 
non-saturated GAN loss~\cite{goodfellow2014generative}, the Wassterstein GAN loss~\cite{arjovsky2017wasserstein,gulrajani2017improved}, the least-square GAN loss~\cite{mao2017least}, the hinge GAN loss~\cite{lim2017geometric,zhang2019self}, the f-divergence GAN loss~\cite{nowozin2016f,jolicoeur2020relativistic}, and the relativistic GAN loss~\cite{jolicoeur2019relativistic}. Empirically, the performance of a GAN loss depends on the application as well as the network architecture. As of the time of writing this survey paper, there is no clear consensus on which one is absolutely better.

Here, we give a generic GAN learning objective formulation that subsumes several popular ones. For the discriminator update step, the learning objective is 
\begin{equation}\label{eqn:dis_update}
\max_{\bm{\phi}} \mathbb{E}_{\bm{x} \sim \mathcal{D}} \Big{[} f_D (D(\bm{x};\bm{\phi})\Big{]} +
\mathbb{E}_{\bm{z} \sim \mathcal{Z}} \Big{[} f_G (D(G(\bm{z};\bm{\theta});\bm{\phi})\Big{]},
\end{equation}
where $f_D$ and $f_G$ are the output layers that transform the results computed by the discriminator $D$ to the classification scores for the real and fake images, respectively. For the generator update step, the learning objective is
\begin{equation}\label{eqn:gen_update}
\min_{\bm{\theta}} \mathbb{E}_{\bm{z} \sim \mathcal{Z}} \Big{[} g_G (D(G(\bm{z};\bm{\theta});\bm{\phi})\Big{]},
\end{equation}
where $g_G$ is the output layer that transforms the result computed by the discriminator to a classification score for the fake image. In Table~\ref{tbl:learning_objective}, we compare $f_D$, $f_G$, and $g_G$ for several popular GAN losses.

\subsection{Training}

Two variants of stochastic gradient descent/ascent (SGD) schemes are commonly used for GAN training: the simultaneous update scheme and the alternating update scheme. Let $V_D(\bm{\theta},\bm{\phi})$ and $V_G(\bm{\theta},\bm{\phi})$ be the objective functions in (\ref{eqn:dis_update}) and (\ref{eqn:gen_update}), respectively. In the simultaneous update, each training iteration contains a discriminator update step and a generator update step given by
\begin{align}
	&\bm{\phi}^{(t+1)} = \bm{\phi}^{(t)} + \alpha_{D} \frac{\partial V_D(\bm{\theta}^{(t)},\bm{\phi}^{(t)})}{\partial \bm{\phi}}\\
	&\bm{\theta}^{(t+1)} = \bm{\theta}^{(t)} - \alpha_{G} \frac{\partial V_G(\bm{\theta}^{(t)},\bm{\phi}^{(t)})}{\partial \bm{\theta}},~\label{eqn:simultaneous_gen}
\end{align} 
where $\alpha_D$ and $\alpha_G$ are the learning rates for the generator and discriminator, respectively. In the alternating update, each training iteration consists of one  discriminator update step followed by a generator update step, given by
\begin{align}
	&\bm{\phi}^{(t+1)} = \bm{\phi}^{(t)} + \alpha_{D} \frac{\partial V_D(\bm{\theta}^{(t)},\bm{\phi}^{(t)})}{\partial \bm{\phi}}~\label{eqn:alternating_dis}\\
	&\bm{\theta}^{(t+1)} = \bm{\theta}^{(t)} - \alpha_{G} \frac{\partial V_G(\bm{\theta}^{(t)},\bm{\phi}^{(t+1)})}{\partial \bm{\theta}}.~\label{eqn:alternating_gen}
\end{align} 
Note that in the alternating update scheme, the generator update~(\ref{eqn:alternating_gen}) utilizes the newly updated discriminator parameters $\bm{\theta}^{(t+1)}$, while, in the simultaneous update~(\ref{eqn:simultaneous_gen}), it does not. These two schemes have their pros and cons. The simultaneous update scheme can be computed more efficiently, as a major part of the computation in the two steps can be shared. In the other hand, the alternating update scheme tends to be more stable as the generator update is computed based on the latest discriminator. Recent GAN works~\cite{gulrajani2017improved,brock2018large,park2019semantic,huang2018multimodal,liu2019few} mostly use the alternating update scheme. Sometimes, the discriminator update~(\ref{eqn:alternating_dis}) is performed several times before computing~(\ref{eqn:alternating_gen})~\cite{brock2018large,gulrajani2017improved}.

Among various SGD algorithms, ADAM~\cite{kingma2014adam}, which is based on adaptive estimates of the first and second order moments, is very popular for training GANs. ADAM has several user-defined parameters. Typically, the first momentum is set to 0, while the second momentum is set to 0.999. The learning rate for the discriminator update is often set to 2 to 4 times larger than the learning rate for the generator update (usually set to 0.0001), which is called the two-time update scales (TTUR)~\cite{heusel2017gans}. We also note that RMSProp~\cite{Tieleman2012} is popular for GAN training~\cite{gulrajani2017improved,karras2017progressive,karras2018style,liu2019few}.


\begin{figure*}[!t]
    \centering
    \includegraphics[width=\textwidth, trim=0 0 0 0, clip]{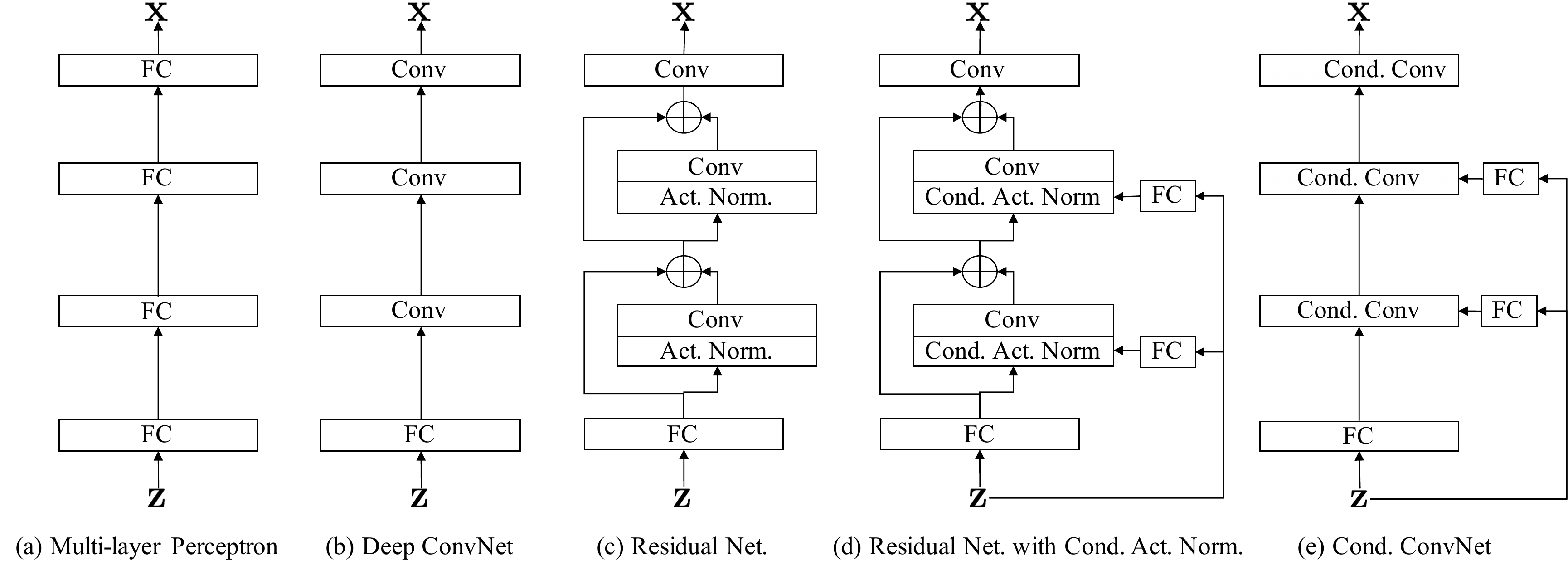}
    \caption{{\bf Generator evolution.} Since the debut of GANs~\cite{goodfellow2014generative}, the generator architecture has continuously evolved. From (a-c), one can observe the change from simple MLPs to deep convolutional and residual networks. Recently, conditional architectures, including conditional activation norms (d) and conditional convolutions (e), have gained popularity as they allow users to have more control on the generation outputs.}
    \label{fig:generator_arch}
\end{figure*}

\subsection{Regularization}
\label{sec:reg}

We review several popular regularization techniques available for countering instability in GAN training.

\mysubsec{Gradient Penalty (GP)} is an auxiliary loss term that penalizes deviation of gradient norm from the desired value~\cite{gulrajani2017improved,mescheder2018training,roth2017stabilizing}. To use GP, one adds it to the objective function for the discriminator update, \ie,~(\ref{eqn:dis_update}). There are several variants of GP. Generally, they can be expressed as 
\begin{equation}
\text{GP-$\delta$} = \mathbb{E}_{\hat{\bm{x}}} \Big{[}\norm{ \nabla D(\hat{\bm{x}}) }_2 - \delta \Big{]}.
\end{equation}
The most common two forms are GP-1~\cite{gulrajani2017improved} and GP-0~\cite{mescheder2018training}.

GP-1 was first introduced by Gulrajani~\etal~\cite{gulrajani2017improved}. It uses an imaginary data distribution 
\begin{equation}
\hat{\bm{x}} = u \bm{x} + (1-u)G(\bm{z}), \quad u\sim \mathcal{U}(0,1)
\end{equation}
where $u$ is a uniform random variable between 0 and 1. Basically, $\hat{\bm{x}}$ is neither real or fake. It is a convex combination of a real sample and a fake sample. The design of the GP-1 is motivated by the property of an optimal $D$ that solves the Wasserstein GAN loss. However, GP-1 is also useful when using other GAN losses. In practice, it has the effect of countering vanishing and exploding gradients occurred during GAN training.

On the other hand, the design of GP-0 is based on the idea of penalizing the discriminator deviating away from the Nash-equilibrium. GP-0 takes a simpler form where they do not use imaginary sample distribution but use the real data distribution, i.e., setting $\hat{\bm{x}}\equiv\bm{x}$. We find the use of GP-0 in several state-of-the-art GAN algorithms~\cite{karras2018style,karras2020analyzing}.

\mysubsec{Spectral Normalization (SN)}~\cite{miyato2018spectral} is an effective regularization technique used in many recent GAN algorithms~\cite{zhang2019self,brock2018large,park2019semantic,saito2020coco}. SN is based on regularizing the spectral norm of the projection operation at each layer of the discriminator, by simply dividing the weight matrix by its largest eigenvalue. Let $W$ be the weight matrix of a layer of the discriminator network. With SN, the true weight that is applied is 
\begin{equation}
W / {\sqrt{\lambda_{max}(W^{T}W)}},    
\end{equation}
where $\lambda_{max}(A)$ extracts the largest eigenvalue from the square matrix $A$. In other word, each project layer has a projection matrix with spectral norm equal to one.

\mysubsec{Feature Matching (FM)} provides a way to encourage the generator to generate images similar to real ones in some sense. Similar to GP, FM is an auxiliary loss. There are two popular implementations: one is batch-based~\cite{salimans2016improved} and the other is instance-based~\cite{larsen2016autoencoding,wang2018high}. Let $D^{i}$ be the $i$-th layer of a discriminator $D$, i.e., $D=D^{d}\circ ...\circ D^2 \circ D^1$. For the batch-based FM loss, it matches the moments of the activations extracted by the real and fake images, respectively. For the $i$-th layer, the loss is
\begin{equation}
\norm{ \mathbb{E}_{\bm{x}\sim \mathcal{D}} [D^{i}\circ ...\circ D^1 (\bm{x})] - 
\mathbb{E}_{\bm{z}\sim \mathcal{Z}} [D^{i}\circ ...\circ D^1 (G(\bm{z}))] }.
\end{equation}
One can apply the FM loss to a subset of layers in the generator and use the weighted sum as the final FM loss. The instance-based FM loss is only applicable to conditional generation models where we have the corresponding real image for a fake image. For the $i$-th layer, the instance-based FM loss is given by
\begin{equation}
\norm{ [D^{i}\circ ...\circ D^1 (\bm{x}_i)] - [D^{i}\circ ...\circ D^1 (G(\bm{z}, \bm{y}_i))] },
\end{equation}
where $\bm{y}_i$ is the control signal for $\bm{x}_i$.

\mysubsec{Perceptual Loss}~\cite{johnson2016perceptual}. Often, when instance-based FM loss is applicable, one can additionally match features extracted from real and fake images using a pretrained network. Such a variant of FM losses is called the perceptual loss~\cite{johnson2016perceptual}.

\mysubsec{Model Average (MA)} can improve the quality of images generated by a GAN. To use MA, we keep two copies of the generator network during training, where one is the original generator with weight $\bm{\theta}$ and the other is the model average generator with weight $\bm{\theta}_{MA}$. At iteration $t$, we update $\bm{\theta}_{MA}$ based on 
\begin{equation}
\bm{\theta}_{MA}^{(t)} = \beta\bm{\theta}^{(t)} + (1-\beta) \bm{\theta}_{MA}^{(t-1)},
\end{equation}
where $\beta$ is a scalar controlling the contribution from the current model weight. 

\subsection{Network Architecture}

Network architectures provide a convenient way to inject inductive biases. Certain network designs often work better than others for a given task. Since the introduction of GANs, we have observed an evolution of the network architecture for both the generator and discriminator.

\mysubsec{Generator Evolution.} In Figure~\ref{fig:generator_arch}, we visualize the evolution of the GAN generator architecture. In the original GAN paper~\cite{goodfellow2014generative}, both the generator and the discriminator are based on the multilayer perceptron (MLP) (Figure~\ref{fig:generator_arch}(a)). As an MLP fails to model the translational invariance property of natural images, its output images are of limited quality. In the DCGAN work~\cite{radford2015unsupervised}, deep convolutional architecture (Figure~\ref{fig:generator_arch}(b)) is used for the GAN generator. As the convolutional architecture is a better fit for modeling image signals, the outputs produced by the DCGAN are often with better quality. Researchers also borrow architecture designs from discriminative modeling tasks. As the residual architecture~\cite{he2016deep} is proven to be effective for training deep networks, several GAN works start to use the residual architecture in their generator design (Figure~\ref{fig:generator_arch}(c))~\cite{arjovsky2017wasserstein,miyato2018spectral}.

A residual block used in modern GAN generators typically consists of a skip connection paired with a series of batch normalization (BN)~\cite{ioffe2015batch}, nonlinearity, and convolution operations. The BN is one type of activation norm (AN), a technique that normalizes the activation values to facilitate training. Other AN variants have also been exploited for the GAN generator, including the instance normalization~\cite{ulyanov2016instance}, the layer normalization~\cite{ba2016layer}, and the group normalization~\cite{wu2018group}. Generally, an activation normalization scheme consists of a whitening step followed by an affine transformation step. Let $\bm{h}_c$ be the output of the whitening step for $\bm{h}$. The final output of the normalization layer is
\begin{equation}
\gamma_c \bm{h}_c + \beta_c,
\end{equation}
where $\gamma_c$ and $\beta_c$ are scalars used to shift the post-normalization activation values. They are constants learned during training.

For many applications, it is required to have some way to control the output produced by a generator. This desire has motivated various conditional generator architectures (Figure~\ref{fig:generator_arch}(d)) for the GAN generator\cite{brock2018large,huang2018multimodal,park2019semantic}. The most common approach is to use the conditional AN. In a conditional AN, both $\gamma_c$ and $\beta_c$ are data dependent. Often, one employs a separate network to map input control signals to the target $\gamma_c$ and $\beta_c$ values. Another way to achieve such controllability is to use hyper-networks; Basically, using an auxiliary network to produce weights for the main network. For example, we can have a convolutional layer where the filter weights are generated by a separate network. We often call such a scheme conditional convolutions (Figure~\ref{fig:generator_arch}(e)), and it has been used for several state-of-the-art GAN generators~\cite{karras2020analyzing,wang2019few}.

\mysubsec{Discriminator Evolution.} GAN discriminators have also undergone an evolution. However, the change has mostly been on moving from the MLP to deep convolutional and residual architectures. As the discriminator is solving a classification task, new breakthroughs in architecture design for image classification tasks could influence future GAN discriminator designs.

\begin{figure}[!t]
    \centering
    \includegraphics[width=\columnwidth, trim=0 0 0 0, clip]{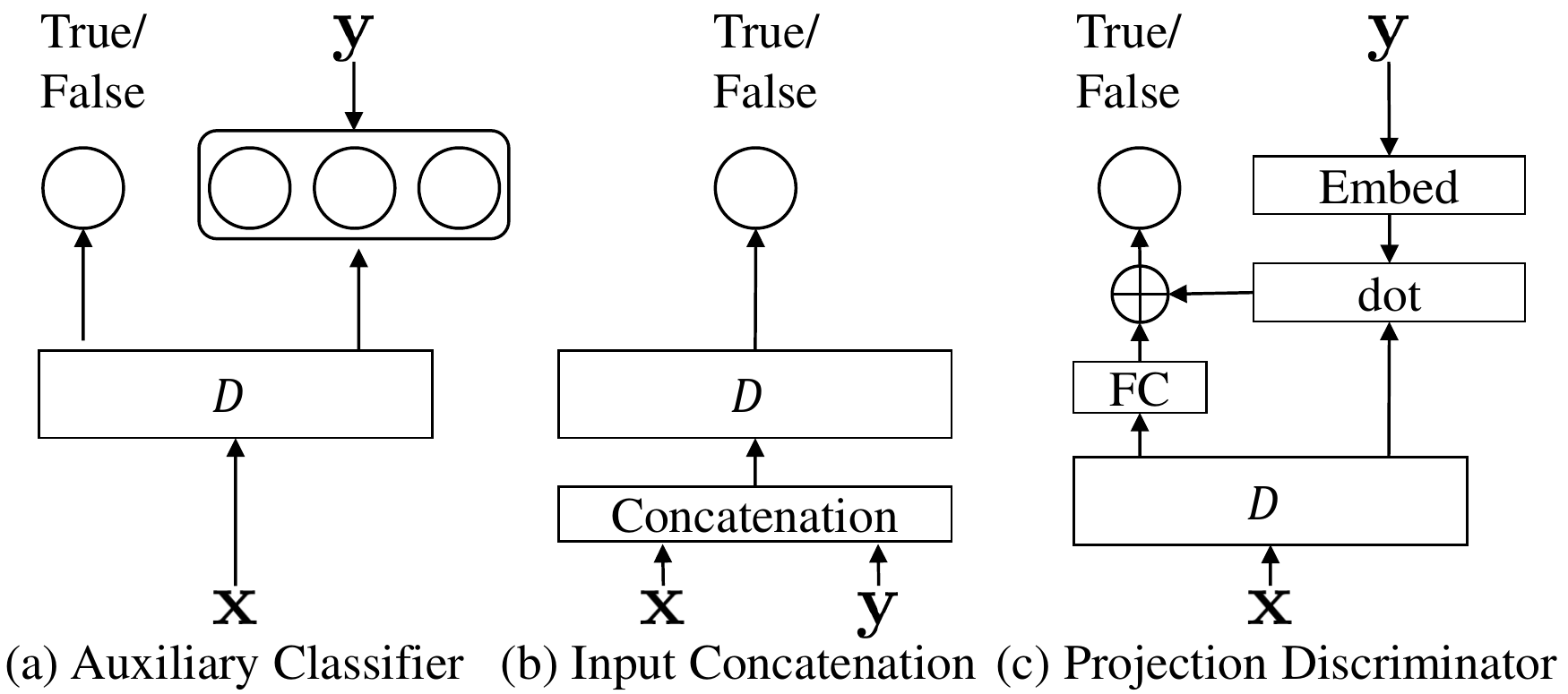}
    \caption{{\bf Conditional discriminator architectures.} There are several ways to leverage the user input signal $\bm{y}$ in the GAN discriminator. (a) \textit{Auxiliary classifier}~\cite{odena2016conditional}. In this design, the discriminator is asked to predict the ground truth label for the real image. 
    (b) \textit{Concatenation}~\cite{isola2017image}. In this design, the discriminator learns to reason whether the input is real by learning a joint feature embedding of image and label. 
    (c) \textit{Projection discriminator}~\cite{miyato2018cgans}. In this design, the discriminator computes an image embedding and correlates it with the label embedding (through the dot product) to determine whether the input is real or fake.}
    \label{fig:discriminator_arch}
\end{figure}

\mysubsec{Conditional Discriminator Architecture.} There are several effective architectures for utilizing control signals (conditional inputs $\bm{y}$) in the GAN discriminator to achieve better image generation quality, as visualized in Figure~\ref{fig:discriminator_arch}. This includes the auxiliary classifier (AC)~\cite{odena2016conditional}, input concatenation (IC)~\cite{isola2017image}, and the projection discriminator (PD)~\cite{miyato2018cgans}. The AC and PD are mostly used for category-conditional image generation tasks, while the PD is common for image-to-image translation tasks.

\mysubsec{Neural Architecture Search.} As neural architecture search has become a popular topic for various recognition tasks, efforts have been made in trying to automatically find a performant architecture for GANs~\cite{gong2019autogan}.

While the current and previous sections have focused on introducing the GAN mechanism and various algorithms used to train them, the following sections focus on various applications of GANs in generating images and videos.

\section{Image Translation}\label{sec:supervised}
This section discusses the application of GANs to image-to-image translation, which aims to map an image from one domain to a corresponding image in a different domain, \eg, sketch to shoes, label maps to photos, summer to winter. The problem can be studied in a supervised setting, where example pairs of corresponding images are available, or an unsupervised setting, where such training data is unavailable and we only have two independent sets of images. In the following subsections, we will discuss recent progress in both settings.

\subsection{Supervised Image Translation}

\begin{figure*}[!t]
	\begin{center}
		\includegraphics[width=\linewidth]{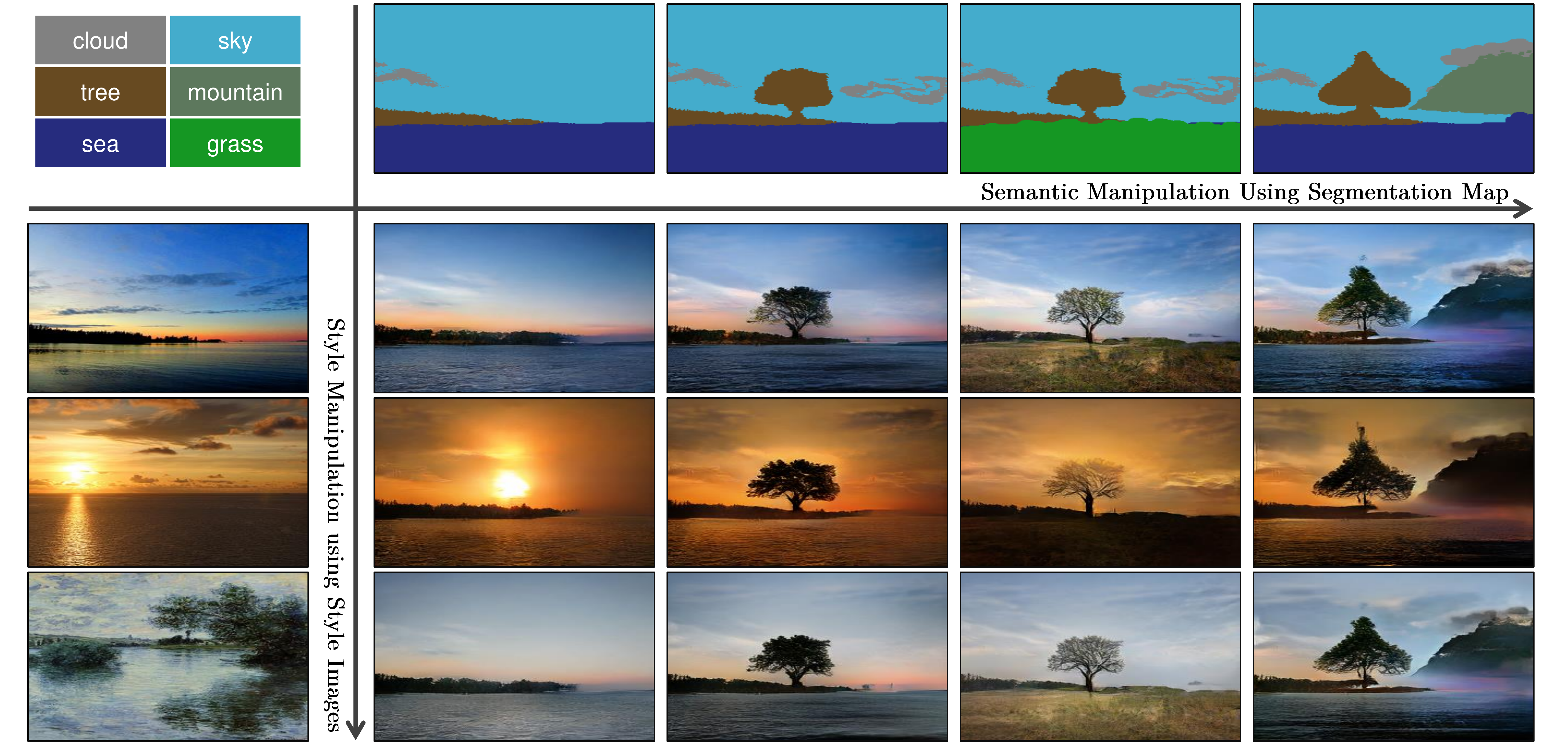}
	\end{center}
	\caption{\textbf{Image translation examples of SPADE~\cite{park2019semantic}}, which converts semantic label maps into photorealistic natural scenes. The style of the output image can also be controlled by by a reference image (the leftmost column). Images are from Park \etal~\cite{park2019semantic}. }
	\label{fig:spade}
\end{figure*}

Isola~\etal~\cite{isola2017image} proposed the pix2pix framework as a general-purpose solution to image-to-image translation in the supervised setting. The training objective of pix2pix combines conditional GANs with the pixel-wise \(\ell_1\) loss between the generated image and the ground truth. One notable design choice of pix2pix is the use of patch-wise discriminators (PatchGAN), which attempts to discriminate each local image patch rather than the whole image. This design incorporates the prior knowledge that the underlying image translation function we want to learn is local, assuming independence between pixels that are far away. In other words, the translation mostly involves style or texture change. It significantly alleviates the burden of the discriminator because it requires much less model capacity to discriminate local patches than whole images.

One important limitation of pix2pix is that its translation function is restricted to be one-to-one. However, many of the mappings we aim to learn are one-to-many in nature. In other words, the distribution of possible outputs is multimodal. For example, one can imagine many shoes in different colors and styles that correspond to the same sketch of a shoe. Naively injecting a Gaussian noise latent code to the generator does not lead to many variations, since the generator is free to ignore that latent code. BicycleGAN~\cite{zhu2017toward} explores approaches to encourage the generator to make use of the latent code to represent output variations, including applying a KL divergence loss to the encoded latent code, and reconstructing the sampled latent code from the generated image. Other strategies to encourage diversity include using different generators to capture different output modes~\cite{ghosh2018multi}, replacing the reconstruction loss with maximum likelihood objective~\cite{lee2019harmonizing,li2019diverse}, and directly encouraging the distance between output images generated from different latent codes to be large~\cite{yang2019diversity,liu2019normalized,mao2019mode}.

Besides, the quality of image-to-image translation has been significantly improved by some recent works~\cite{wang2018high,park2019semantic,liu2019learning,lee2020maskgan,tang2020edge,zheng2020example}. In particular, pix2pixHD~\cite{wang2018high} is able to generate high-resolution images with a coarse-to-fine generator and a multi-scale discriminator. SPADE~\cite{park2019semantic} further improves the image quality with a spatially-adaptive normalization layer. SPADE, in addition, allows a style image input for better control the desired look of the output image. Some examples of SPADE are shown in Figure~\ref{fig:spade}.

\subsection{Unsupervised Image Translation}
For many tasks, paired training images are very difficult to obtain~\cite{zhu2017unpaired,kim2017learning,yi2017dualgan,liu2016unsupervised,benaim2017one,huang2018multimodal,choi2017stargan,lee2018diverse}. Unsupervised learning of mappings between corresponding images in two domains is a much harder problem but has wider applications than the supervised setting. CycleGAN~\cite{zhu2017unpaired} simultaneously learns mappings in both directions and employs a cycle consistency loss to enforce that if an image is translated to the other domain and translated back to the original domain, the output should be close to the original image. UNIT~\cite{liu2016unsupervised} makes a shared latent space assumption~\cite{liu2016coupled} that a pair of corresponding images can be mapped to the same latent code in a shared latent space. It is shown that shared-latent space implies cycle consistency and imposes a stronger regularization. DistanceGAN~\cite{benaim2017one} encourages the mapping to preserve the distance between any pair of images before and after translation. While the methods above need to train a different model for each pair of image domains, StarGAN~\cite{choi2017stargan} is able to translate images across multiple domains using only a single model.

In many unsupervised image translation tasks (\eg, horses to zebras, dogs to cats), the two image domains mainly differ in the foreground objects, and the background distribution is very similar. Ideally, the model should only modify the foreground objects and leave the background region untouched. Some work~\cite{chen2018attention,mejjati2018unsupervised,yang2019show} employs spatial attention to detect and change the foreground region without influencing the background. InstaGAN~\cite{mo2019instagan} further allows the shape of the foreground objects to be changed.

\begin{figure}[!t]
	\begin{center}
		\includegraphics[width=\columnwidth]{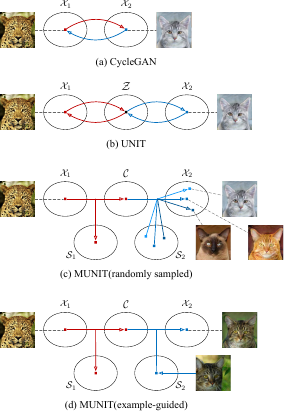}
	\end{center}
	\caption{\textbf{Comparisons among unsupervised image translation methods (CycleGAN~\cite{zhu2017unpaired}, UNIT~\cite{liu2016unsupervised}, and MUNIT~\cite{huang2018multimodal}).} $\mathcal{X}_1$ and $\mathcal{X}_2$ are two different image domains (dogs and cats in this example). (a) CycleGAN enforces the learned mappings to be inverses of each other. (b) UNIT auto-encodes images in both domains to a common latent space $\mathcal{Z}$. Both CycleGAN and UNIT can only perform unimodal translation. (c) MUNIT decomposes the latent space into a shared content space $\mathcal{C}$ and unshared style spaces $\mathcal{S}_1, \mathcal{S}_2$. Diverse outputs can be obtained by sampling different style codes from the target style space. (d) The style of the translation output can also be controlled by a guiding image in the target domain.}
	\label{fig:unsupervised_translation}
\end{figure}

The early work mentioned above focuses on unimodal translation. On the other hand, recent advances~\cite{huang2018multimodal,lee2018diverse,almahairi2018augmented,gonzalez2018image,mao2019mode,ma2019exemplar} have made it possible to perform multimodal translation, generating diverse output images given the same input. For example, MUNIT~\cite{huang2018multimodal} assumes that images can be encoded into two disentangled latent spaces: a domain-invariant content space that captures the information that should be preserved during translation, and a domain-specific style space that represents the variations that are not specified by the input image. To generate diverse translation results, we can recombine the content code of the input image with different style codes sampled from the style space of the target domain. Figure~\ref{fig:unsupervised_translation} compares MUNIT with existing unimodal translation methods including CycleGAN and UNIT.
The disentangled latent space not only enables multimodal translation, but also allows example-guided translation in which the generator recombines the domain-invariant content of an image from the source domain and the domain-specific style of an image from the target domain. The idea of using a guiding style image has also been applied to the supervised setting~\cite{park2019semantic,pix2pixSC2019,zhang2020cross}.

Although paired example images are not needed in the unsupervised setting, most existing methods still require access to a large number of unpaired example images in both source and target domains. Some works seek to reduce the number of training examples without much loss of performance. Benaim and Wolf~\cite{benaim2018one} focus on the situation where there are many images in the target domain but only a single image in the source domain. The work of Cohen and Wolf~\cite{cohen2019bidirectional} enables translation in the opposite direction where the source domain has many images but the target domain has only one. The above setting assumes the source and target domain images, whether there are many or few, are available during training. Liu~\etal~\cite{liu2019few} proposed FUNIT to address a different situation where there are many source domain images that are available during training, but few target domain images that are available only at test time. The target domain images are used to guide translation similar to the example-guided translation procedure in MUNIT. Saito~\etal~\cite{saito2020coco} proposed a content-conditioned style encoder to better preserve the domain-invariant content of the input image. However, the above scenario~\cite{liu2019few,saito2020coco} still assumes access to the domain labels of the training images. Some recent work aims to reduce the need for such supervision by using few~\cite{wang2020semi} or even no~\cite{baek2020rethinking} domain labels. Very recently, some works~\cite{lin2020tuigan,benaim2020structural,park2020contrastive} are able to achieve image translation even when each domain only has a single image, inspired by recent advances that can train GANs on a single image~\cite{shaham2019singan}.

Despite the empirical successes, the problem of unsupervised image-to-image translation is inherently ill-posed, even with constraints such as cycle consistency or shared latent space. Specifically, there exist infinitely many mappings that satisfy those constraints~\cite{galanti2018role,yang2018esther,de2019optimal}, yet most of them are not semantically meaningful. How do current methods successfully find the meaningful mapping in practice? Galanti~\etal~\cite{galanti2018role} assume that the meaningful mapping is of minimal complexity and the popular generator architectures are not expressive enough to represent mappings that are highly complex. Bezenac~\etal~\cite{de2019optimal} further argue that the popular architectures are implicitly biased towards mappings that produce minimal changes to the input, which are usually semantically meaningful. In summary, the training objectives of unsupervised image translation alone cannot guarantee that the model can find semantically meaningful mappings and the inductive bias of generator architectures plays an important role.

\section{Image Processing}
\label{sec:image_processing}

GAN's strength in generating realistic images makes it ideal for solving various image processing problems, especially for those where the perceptual quality of image outputs is the primary evaluation criteria. This section will discuss some prominent GAN-based methods for several key image processing problems, including image restoration and enhancement (super-resolution, denoising, deblurring, compression artifacts removal) and image inpainting.

\subsection{Image Restoration and Enhancement}

\begin{figure}[!t]
  \begin{center}
  \includegraphics[width=0.9\linewidth]{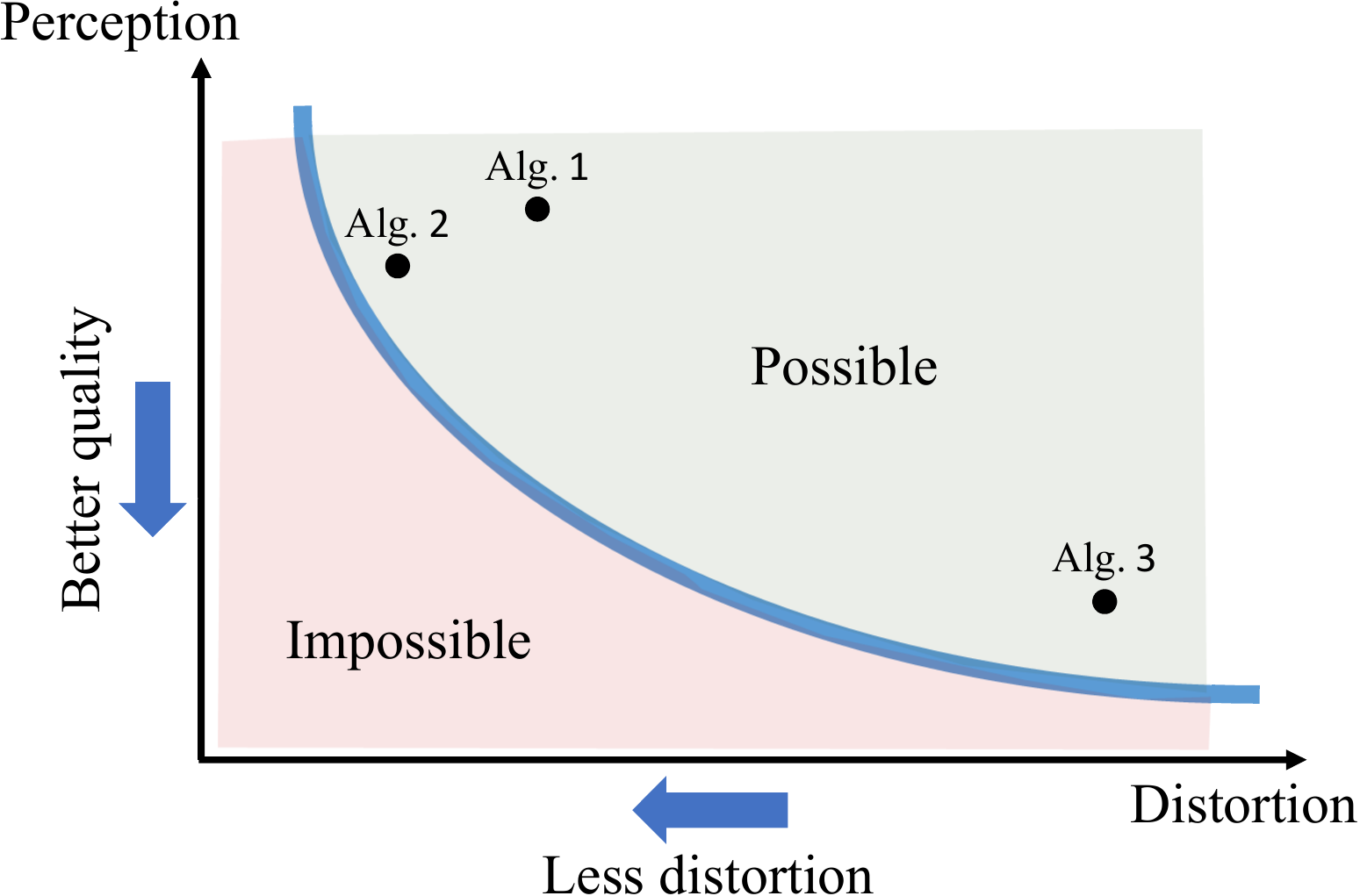}
  \end{center}
  \caption{\textbf{Perception-distortion tradeoff~\cite{blau2018perception}.} Distortion metrics, including the MSE, PSNR, and SSIM, measure the similarity between the ground truth image and the restored images. Perceptual quality metrics, including NR~\cite{ma2017learning}, measure the distribution distance between the recovered image distribution and the target image distribution. Blau~\etal~\cite{blau2018perception} show that an image restoration algorithm can be characterized by the distortion and perceptual quality tradeoff curve. The plot is from Blau~\etal~\cite{blau2018perception}.}
\label{fig:perc_dist_tradeoff}
\end{figure}

The traditional way of evaluating algorithms for image restoration and enhancement tasks is to measure the \textit{distortion}, the difference between the ground truth images and restored images using metrics like the mean square error (MSE), the peak signal-to-noise ratio (PSNR), and the structural similarity index (SSIM). Recently, metrics for measuring \textit{perceptual quality}, such as the no-reference (NR) metric~\cite{ma2017learning}, have been proposed, as the visual quality is arguably the most important factor for the usability of an algorithm. Blau \etal~\cite{blau2018perception} proposed the perception-distortion tradeoff~\cite{blau2018perception}, which states that an image restoration algorithm can potentially improve only in terms of its distortion or in terms of its perceptual quality, as shown in the Figure~\ref{fig:perc_dist_tradeoff}. Blau \etal~\cite{blau2018perception} further demonstrate that GANs provide a principled way to approach the perception-distortion bound.

\mysubsec{Image super-resolution (SR)} aims at estimating a high-resolution (HR) image from its low-resolution (LR) counterpart. Deep learning has enabled faster and more accurate super-resolution methods, including SRCNN~\cite{dong2015image}, FSRCNN~\cite{dong2016accelerating}, ESPCN~\cite{shi2016real}, VDSR~\cite{kim2016accurate}, SRResNet~\cite{ledig2017photo}, EDSR~\cite{lim2017enhanced}, SRDenseNet~\cite{tong2017image}, MemNet~\cite{tai2017memnet}, RDN~\cite{zhang2018residual}, WDSR~\cite{yu2019wide}, and many others. However, the above super-resolution approaches focus on improving the distortion metrics and pay little to no attention to the perceptual quality metrics. As a result, they tend to predict over-smoothed outputs and fail to synthesize finer high-frequency details.

\begin{figure}[!t]
  \begin{center}
  \includegraphics[width=0.9\columnwidth,trim={0 0.5cm 7cm 0},clip]{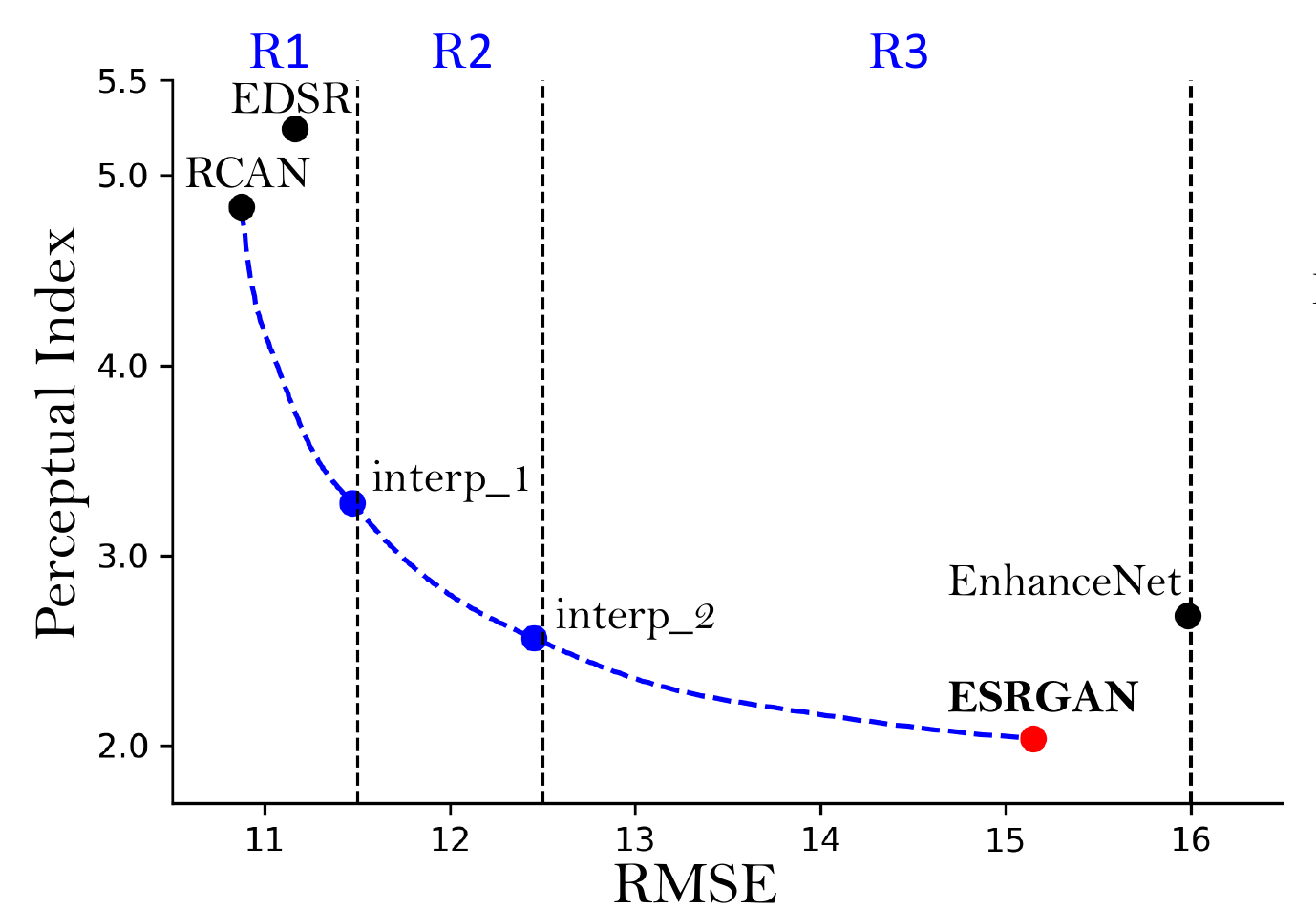}
  \end{center}
  \caption{\textbf{The perception-distortion curve of the ESRGAN~\cite{wang2018esrgan} on PIRM self-validation dataset~\cite{blau20182018}.} The curve also compares the ESRGAN with the EnhanceNet~\cite{sajjadi2017enhancenet}, the RCAN~\cite{zhang2018image}, and the EDSR~\cite{lim2017enhanced}. The curve is from Wang~\etal~\cite{wang2018esrgan}.}
\label{fig:esrgan_perc_dist}
\vspace{2mm}
  \begin{center}
  \includegraphics[width=\linewidth]{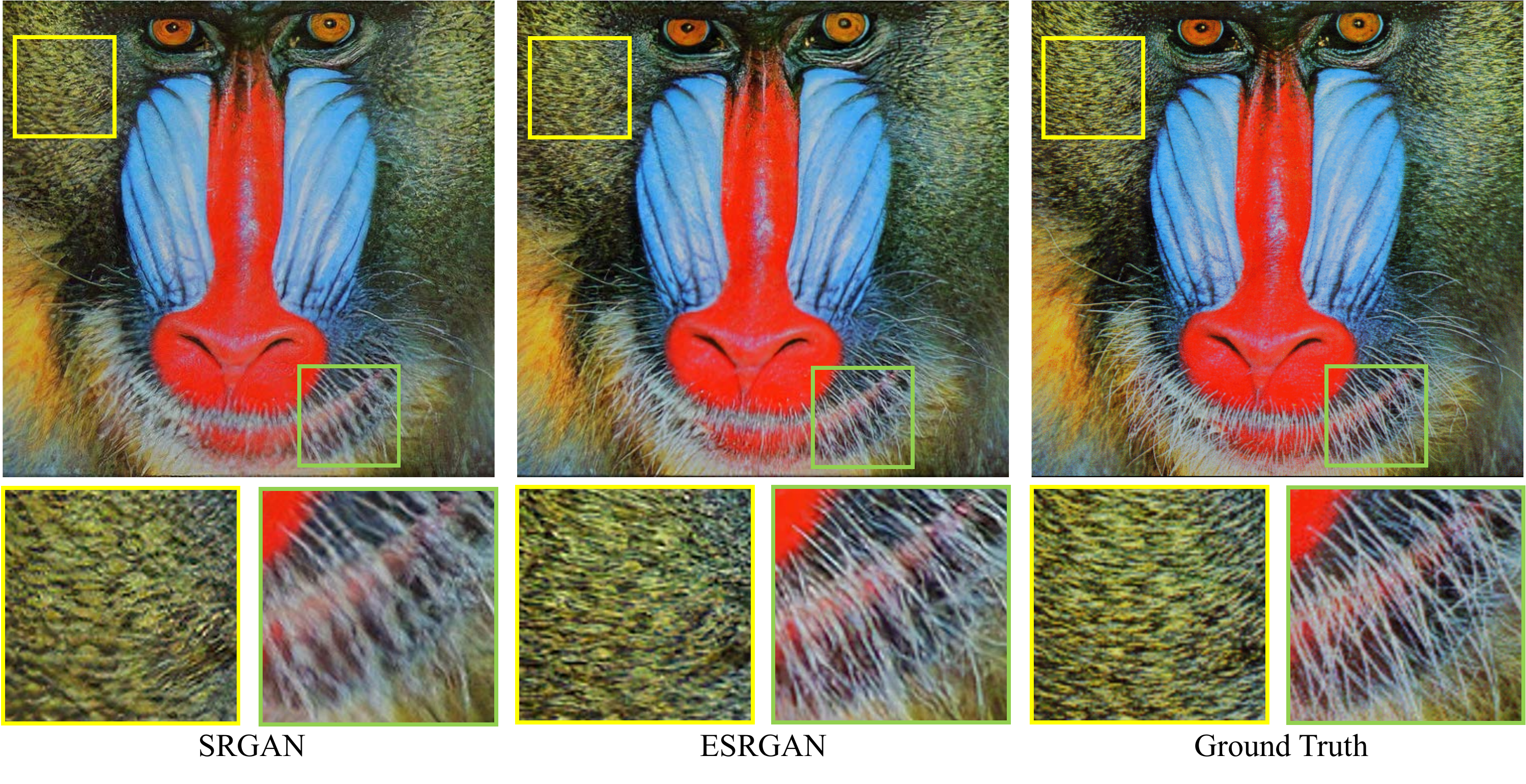}
  \end{center}
  \caption{\textbf{Visual comparison between the ESRGAN~\cite{wang2018esrgan} and the SRGAN~\cite{ledig2017photo}.} Images are from Wang~\etal~\cite{wang2018esrgan}.}
\label{fig:esrgan_visual_example}
\end{figure}

Recent image super-resolution algorithms improve the perceptual quality of outputs by leveraging GANs. The SRGAN~\cite{ledig2017photo} is the first of its kind and can generate photo-realistic images with \(4\times\) or higher upscaling factors. The quality of the SRGAN~\cite{ledig2017photo} outputs is mainly measured by the mean opinion score (MOS) over 26 raters. To enhance the visual quality further, Wang \etal~\cite{wang2018esrgan} revisit the design of the three key components in the SRGAN: the network architecture, the GAN loss, and the perceptual loss. They propose the Enhanced SRGAN (ESRGAN), which achieves consistently better visual quality with more realistic and natural textures than the competing methods, as shown in Figure~\ref{fig:esrgan_perc_dist} and Figure~\ref{fig:esrgan_visual_example}. The ESRGAN is the winner of the 2018 Perceptual Image Restoration and Manipulation challenge (PIRM)~\cite{blau20182018} (region 3 in Figure~\ref{fig:esrgan_perc_dist}). Other GAN-based image super-resolution methods and practices can be found in the 2018 PIRM challenge report~\cite{blau20182018}.

The above image super-resolution algorithms all operate in the supervised setting where they assume corresponding low-resolution and high-resolution pairs in the training dataset. Typically, they create such a training dataset by downsampling the ground truth high-resolution images. However, the downsampled high-resolution images are very different from the low-resolution images captured by a real sensor, which often contain noise and other distortion. As a result, these super-resolution algorithms are not directly applicable to upsample low-resolution images captured in the wild. Several methods have addressed the issue by studying image super-resolution in the unsupervised setting where they only assume a dataset of low-resolution images captured by a sensor and a dataset of high-resolution images. Recently, Maeda~\cite{maeda2020unpaired} proposes a GAN-based image super-resolution algorithm operates in the unsupervised setting for bridging the gap.

\mysubsec{Image denoising} aims at removing noise from noisy images. The task is challenging since the noise distribution is usually unknown. This setting is also referred to as blind image denoising. DnCNN~\cite{zhang2017beyond} is one of the first approaches using feed-forward convolutional neural networks for image denoising. However, DnCNN~\cite{zhang2017beyond} requires knowing the noise distribution in the noisy image and hence has limited applicability. To tackle blind image denoising, Chen \etal~\cite{chen2018image} proposed the GAN-CNN-based Blind Denoiser (GCBD), which consists of 1) a GAN trained to estimate the noise distribution over the input noisy images to generate noise samples, and 2) a deep CNN that learns to denoise on generated noisy images. The GAN training criterion of GCBD~\cite{chen2018image} is based on Wasserstein GAN~\cite{arjovsky2017wasserstein}, and the generator network is based on DCGAN~\cite{radford2015unsupervised}.

\begin{figure}[!t]
	\begin{center}
	\begin{tabular}{ccc}	
		\includegraphics[width=0.28\linewidth]{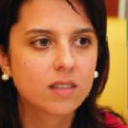}&
		\includegraphics[width=0.28\linewidth]{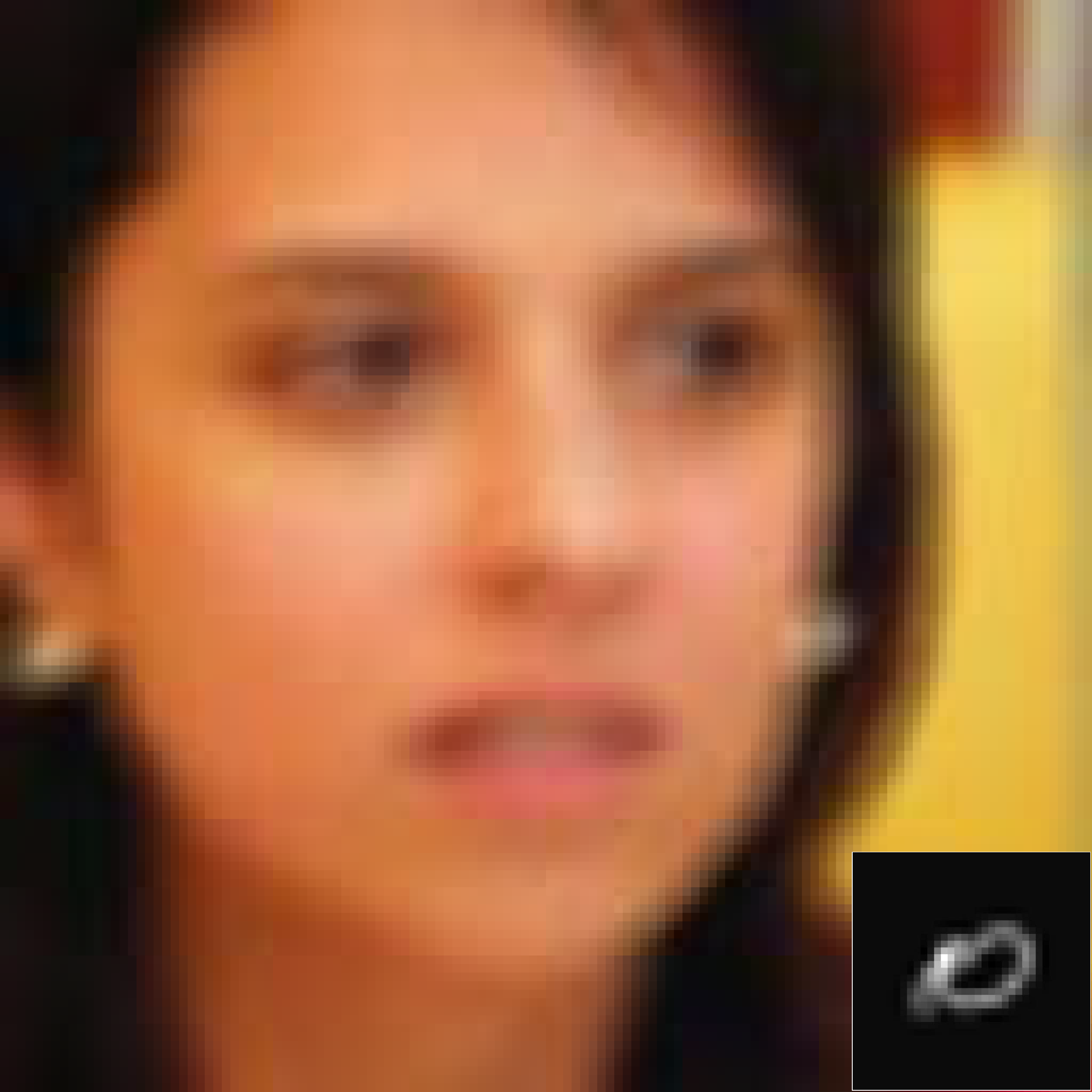}&
		\includegraphics[width=0.28\linewidth]{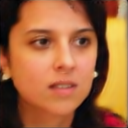}  \\
		\includegraphics[width=0.28\linewidth]{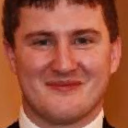}&
		\includegraphics[width=0.28\linewidth]{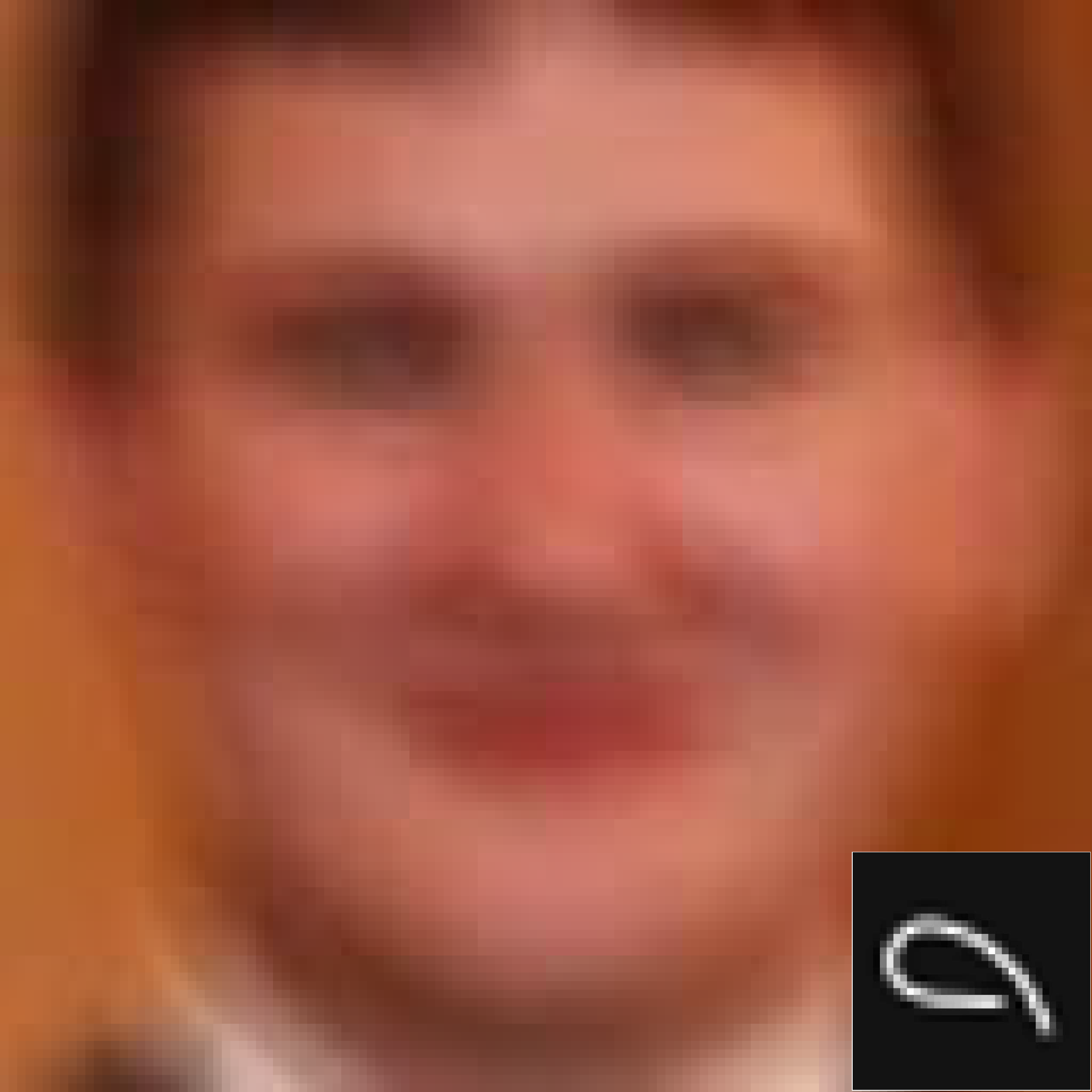}&
		\includegraphics[width=0.28\linewidth]{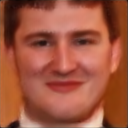} \\
        Ground truth & Blurry inputs & Deblurred results\\
    \end{tabular}	
	\end{center}
	\vspace{-2mm}
	\caption{\textbf{Face deblurring results with GANs~\cite{shen2018deep}.} Images are from Shen~\etal~\cite{shen2018deep}.}
	\label{fig:face_deblur}
	\vspace{-2mm}
\end{figure}

\mysubsec{Image deblurring} sharpens blurry images, which result from motion blur, out-of-focus, and possibly other causes. DeblurGAN~\cite{kupyn2018deblurgan} trains an image motion deblurring network using Wasserstein GAN~\cite{arjovsky2017wasserstein} with the GP-1 loss and the perceptual loss (See Section~\ref{sec:learning}). Shen \etal~\cite{shen2018deep} use a similar approach to deblur face image by using GAN and perceptual loss and incrementally training the deblurring network. Visual examples are shown in Figure~\ref{fig:face_deblur}.

\begin{figure}
    \tabcolsep=1pt\relax
\footnotesize
\centering
    \begin{tabular}{llrllrllr}
        \multicolumn{3}{c}{%
            \includegraphics[width=.3\linewidth]{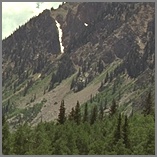}}&
        \multicolumn{3}{c}{%
            \includegraphics[width=.3\linewidth]{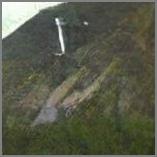}}&
        \multicolumn{3}{c}{%
            \includegraphics[width=.3\linewidth]{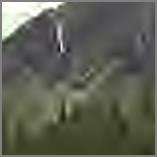}}\\[-1pt]

        \multicolumn{3}{l}{Original}&
        GAN~\cite{Agustsson_2019_ICCV}: & \multicolumn{2}{l}{1567 B, 1x}&
        JP2K: & 3138 B, 2x & \\[2pt]
        \multicolumn{3}{c}{%
            \includegraphics[width=.3\linewidth]{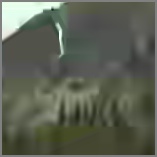}}&
        \multicolumn{3}{c}{%
            \includegraphics[width=.3\linewidth]{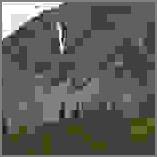}}&
        \multicolumn{3}{c}{%
            \includegraphics[width=.3\linewidth]{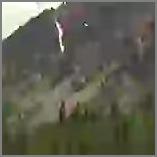}}\\[-1pt]
        BPG: & 3573 B, 1.2x & &
        JPEG: & 13959 B, 7.9x & &
        WebP: & 9437 B, 5x & \\[2pt]
    \end{tabular}
    
\caption{\textbf{Image compression with GANs~\cite{Agustsson_2019_ICCV}.} Comparing a GAN-based approach~\cite{Agustsson_2019_ICCV} for image compression to those obtained by the off-the-shelf codecs. Even with fewer than half the number of bytes, GAN-based compression~\cite{Agustsson_2019_ICCV} produces more realistic visual results. Images are from from Agustsson \etal~\cite{Agustsson_2019_ICCV}.}
\label{fig:image_compression_artifacts_removal}
\end{figure}

\mysubsec{Lossy image compression} algorithms (\eg, JPEG, JPEG2000, BPG, and WebP) can efficiently reduce image sizes but introduce visual artifacts in compressed images when the compression ratio is high. Deep neural networks have been widely explored for removing the introduced artifacts~\cite{galteri2017deep, Agustsson_2019_ICCV, tschannen2018deep}. Galteri \etal~\cite{galteri2017deep} show that a residual network trained with a GAN loss is able to produce images with more photorealistic details than MSE or SSIM-based objectives for the removal of image compression artifacts. Tschannen \etal~\cite{tschannen2018deep} further proposed distribution-preserving lossy compression by using a new combination of Wasserstein GAN and Wasserstein autoencoder~\cite{tolstikhin2018wasserstein}. More recently, Agustsson \etal~\cite{Agustsson_2019_ICCV} built an extreme image compression system by using unconditional and conditional GANs, outperforming all other codecs in the low bit-rate setting. Some compression visual examples~\cite{Agustsson_2019_ICCV} are shown in Figure~\ref{fig:image_compression_artifacts_removal}.

\subsection{Image Inpainting}

Image inpainting aims at filling missing pixels in an image such that the result is visually realistic and semantically correct. Image inpainting algorithms can be used to remove distracting objects or retouch undesired regions in photos and can be further extended to other tasks, including image un-cropping, rotation, stitching, re-targeting, re-composition, compression, super-resolution, harmonization, and more.

Traditionally patch-based approaches, such as the PatchMatch~\cite{barnes2009patchmatch}, copy background patches according to the low-level feature matching (\eg, euclidean distance on pixel RGB values) and paste them into the missing regions. These approaches can synthesize plausible stationary textures but fail at non-stationary image regions such as faces, objects, and complicated scenes. Recently, deep learning and GAN-based approaches~\cite{yu2018generative, liu2018image, yu2019free, nazeri2019edgeconnect, xiong2019foreground, wang2018image, zheng2019pluralistic, jo2019sc, zeng2019learning, iizuka2017globally} open a new direction for image inpainting using deep neural networks learned on large-scale data in an end-to-end fashion. Comparing to PatchMatch, these methods are more scalable and can leverage large-scale data.

The context encoder approach (CE)~\cite{pathak2016context} is one of the first in using a GAN generator to predict the missing regions and is trained with the \(\ell_2\) pixel-wise reconstruction loss and a GAN loss. Iizuka \etal~\cite{iizuka2017globally} further improve the GAN-based inpainting framework by using both global and local GAN discriminators, with the global one operating on the entire image and the local one operating on only the patch in the hole. We note that the post-processing techniques such as image blending are still required in these GAN-based approaches~\cite{pathak2016context, iizuka2017globally} to reduce visual artifacts near hole boundaries.

\begin{figure}[!t]
\begin{center}
    \includegraphics[width=0.48\linewidth]{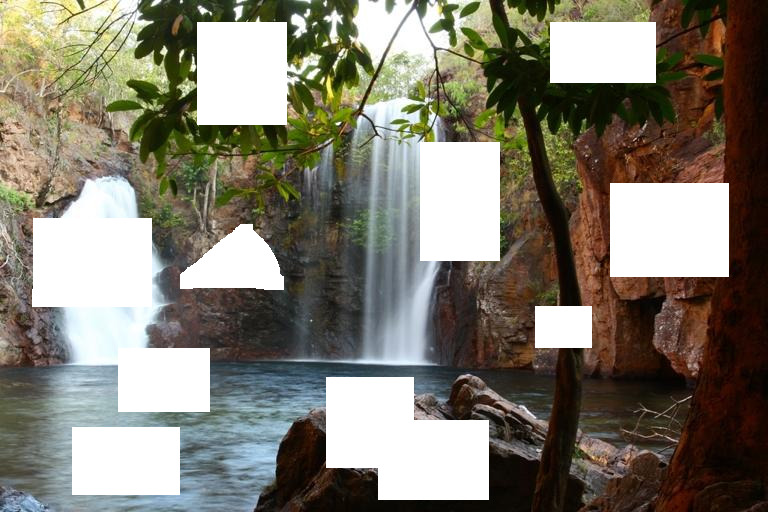}
    \includegraphics[width=0.48\linewidth]{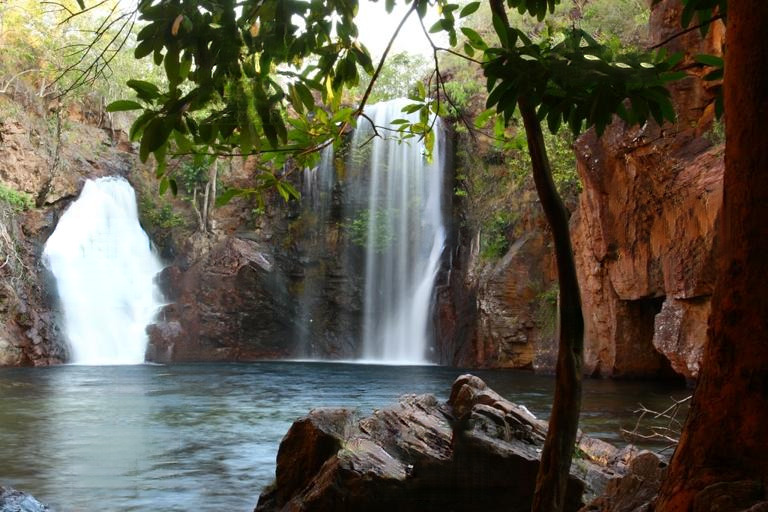}
    \\\vspace{1mm}
    \includegraphics[width=0.235\linewidth]{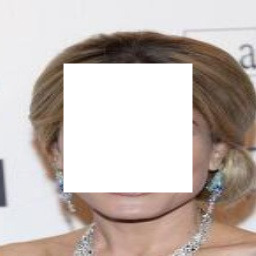}
    \includegraphics[width=0.235\linewidth]{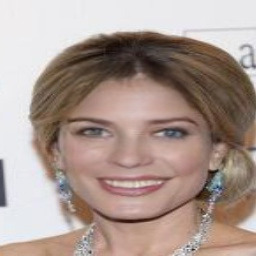}
    \includegraphics[width=0.235\linewidth]{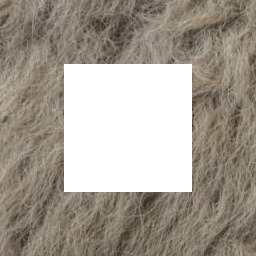}
    \includegraphics[width=0.235\linewidth]{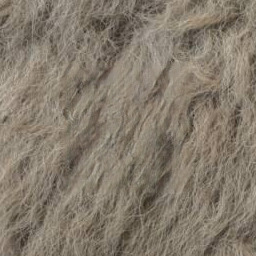}
\caption{\textbf{Image inpainting results uisng the DeepFill~\cite{yu2018generative}.}  Missing regions are shown in white. In each pair, the left is the input image, and the right is the direct output of trained GAN without any post-processing. Images are from Yu~\etal~\cite{yu2018generative}.}
\label{fig:deepfill}
\end{center}
\end{figure}

Yu \etal~\cite{yu2018generative} proposed DeepFill, a GAN framework for end-to-end image inpainting without any post-processing step, which leverages a stacked network, consisting of a coarse network and a refinement network, to ensure the color and texture consistency between the in-filled regions and their surrounding. Moreover, as convolutions are local operators and less effective in capturing long-range spatial dependencies, the contextual attention layer~\cite{yu2018generative} is introduced and integrated into the DeepFill to borrow information from distant spatial locations explicitly. Visual examples of the DeepFill~\cite{yu2018generative} are shown in Figure~\ref{fig:deepfill}.

\begin{figure}[!t]
\begin{center}
    \includegraphics[width=\linewidth]{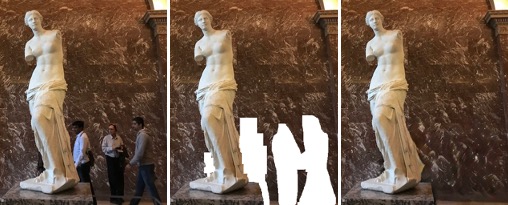}
    \caption{\textbf{Free-form image inpainting results using the DeepFillV2~\cite{yu2019free}.} From left to right, we have the ground truth image, the free-form mask, and the DeepFillV2 inpainting result. Original images are from Yu~\etal~\cite{yu2019free}.}
    \label{fig:deepfillv2}
\end{center}\vspace{2mm}
\begin{center}
    \includegraphics[width=\columnwidth]{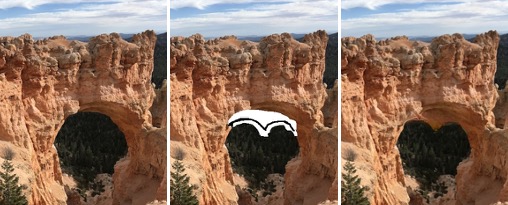}
    \caption{\textbf{User-guided image inpainting results using the DeepFillV2~\cite{yu2019free}.} From left to right, we have the ground truth image, the mask with user-provided edge guidance, and the DeepFillV2 inpainting result. Images are from Yu~\etal~\cite{yu2019free}.}
    \label{fig:deepfillv2_guided}
\end{center}
\end{figure}

One common issue with the earlier GAN-based inpainting approaches~\cite{pathak2016context, iizuka2017globally, yu2018generative} is that the training is performed with randomly sampled rectangular masks. While allowing easy processing during training, these approaches do not generalize well to free-form masks, irregular masks with arbitrary shapes. To address the issue, Liu \etal~\cite{liu2018image} proposed the partial convolution layer where the convolution is masked and re-normalized to utilize valid pixels only. Yu \etal~\cite{yu2019free} further proposed the gated convolution layer, generalizing the partial convolution by providing a learnable dynamic feature selection mechanism for each channel at each spatial location across all layers. In addition, as free-form masks may appear anywhere in images with any shape, global and local GANs~\cite{iizuka2017globally} designed for a single rectangular mask are not applicable. To address this issue, Yu \etal~\cite{yu2019free} introduced a patch-based GAN loss, SNPatchGAN~\cite{yu2019free}, by applying spectral-normalized discriminator on the dense image patches. Visual examples of the DeepFillV2~\cite{yu2019free} with free-form masks are shown in Figure~\ref{fig:deepfillv2}.


Although capable of handling free-form masks, these inpainting methods perform poorly in reconstructing foreground details. This motivated the design of edge-guided image inpainting methods~\cite{nazeri2019edgeconnect, xiong2019foreground}. These methods decompose inpainting into two stages. The first stage predicts edges or contours of foregrounds, and the second stage takes predicted edges to predict the final output. Moreover, for image inpainting, enabling user interactivity is essential as there are many plausible solutions for filling a hole in an image. User-guided inpainting methods~\cite{yu2019free, xiong2019foreground, nazeri2019edgeconnect} have been proposed to provide an option to take additional user inputs, for example, sketches, as guidance for image inpainting networks. An example of user-guided image inpainting is shown in Figure~\ref{fig:deepfillv2_guided}.

Finally, we note that the image out-painting or extrapolation tasks are closely related to image inpainting~\cite{teterwak2019boundless, kim2020painting}. They can be also benefited from a GAN formulation.
\section{Video Synthesis}\label{sec:video}

Video synthesis focuses on generating video content instead of static images. Compared with image synthesis, video synthesis needs to ensure the temporal consistency of the output videos. This is usually achieved by using a temporal discriminator~\cite{tulyakov2018mocogan}, flow-warping loss on neighboring frames~\cite{wang2018video}, smoothing the inputs before processing~\cite{chan2019everybody}, or a post-processing step~\cite{lai2018learning}. Each of them might be suitable for a particular task.

Similar to image synthesis, video synthesis can be classified into unconditional and conditional video synthesis. Unconditional video synthesis generates sequences using random noise inputs~\cite{vondrick2016generating, saito2017temporal, tulyakov2018mocogan, clark2019efficient}. Because such a method needs to model all the spatial and temporal content in a video, the generated results are often short or with very constrained motion patterns. For example, MoCoGAN~\cite{tulyakov2018mocogan} decomposes the motion and content parts of the sequence and uses a fixed latent code for the content and a series of latent codes to generate the motion. The synthesized videos are usually up to a few seconds on simple video content, such as facial motion.

On the other hand, conditional video synthesis generates videos conditioning on input content. A common category is future frame prediction~\cite{srivastava2015unsupervised,kalchbrenner2016video, finn2016unsupervised,mathieu2015deep,lotter2016deep,xue2016visual,walker2016uncertain,walker2017pose,denton2017unsupervised,villegas2017decomposing,liang2017dual,lee2018stochastic}, which attempts to predict the next frame of a sequence based on the past frames.
Another common category of conditional video synthesis is conditioning on an input video that shares the same high-level representation. Such a setting is often referred to as the video-to-video synthesis~\cite{wang2018video}. This line of works has shown promising results on various tasks, such as transforming high-level representations to photorealistic videos~\cite{wang2018video}, animating characters with new expressions, or motions~\cite{thies2016face2face, chan2019everybody}, or innovating a new rendering pipeline for graphics engines~\cite{gafni2019vid2game}. Due to its broader impact, we will mainly focus on conditional video synthesis. Particularly, we will focus on its two major domains: \emph{face reenactment} and \emph{pose transfer}.

\subsection{Face Reenactment}
\begin{figure*}[!t]
    \centering
    \includegraphics[width=.9\textwidth]{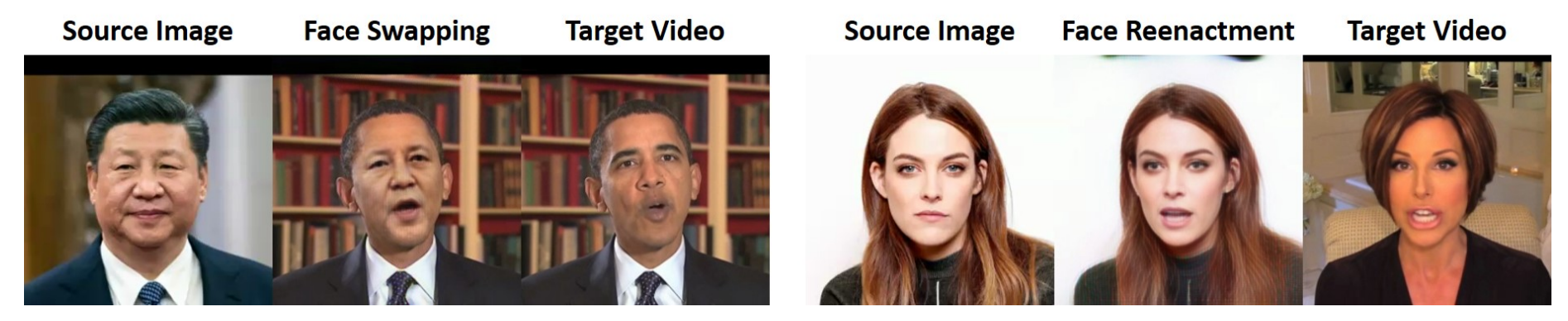}
    \caption{{\bf Face swapping vs.\ reenactment~\cite{nirkin2019fsgan}.} Face swapping focuses on pasting the face region from one subject to another, while face reenactment concerns transferring the expressions and head poses from the target subject to the source image. Images are from Nirkin \etal~\cite{nirkin2019fsgan}.}
    \label{fig:face_swap_vs_reenact}
\end{figure*}

Conditional face video synthesis exists in many forms. The most common forms include \emph{face swapping} and \emph{face reenactment}. Face swapping focuses on pasting the face region from one subject to another, while face reenactment concerns transferring the subject's expressions and head poses.  Figure~\ref{fig:face_swap_vs_reenact} illustrates the difference. Here, we only focus on face reenactment. It has many applications in fields like gaming or film industry, where the characters can be animated by human actors. Based on whether the trained model can only work for a specific person or is universal to all persons, face reenactment can be classified as \emph{subject-specific} or \emph{subject-agnostic} as described below.

\begin{figure*}[!t]
    \centering
    \includegraphics[width=.9\textwidth]{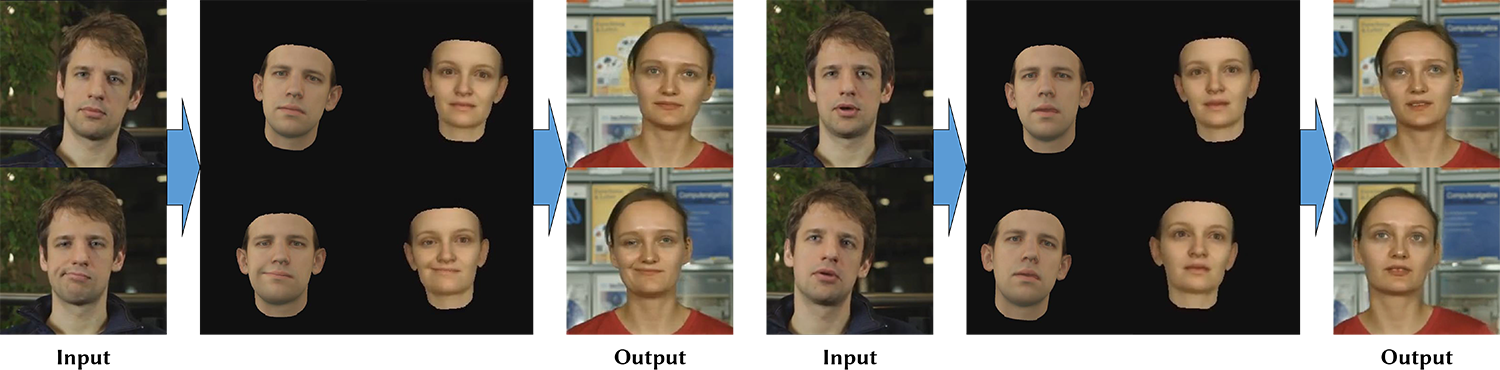}
    \caption{{\bf Face reenactment using 3D face models~\cite{kim2018deep}.} These methods first construct a 3D model for the person to be synthesized, so they can easily animate the model with new expressions. Images are from Kim \etal~\cite{kim2018deep}.}
    \label{fig:face_reenact_3D}
\end{figure*}

\mysubsec{Subject-specific.}
Traditional methods usually build a subject-specific model, which can only synthesize one pre-determined subject by focusing on transferring the expressions without transferring the head movement~\cite{vlasic2005face,thies2015real,thies2016face2face,suwajanakorn2017synthesizing,thies2019deferred}. This line of works usually starts by collecting footage of the target person to be synthesized, either using an RGBD sensor~\cite{thies2015real} or an RGB sensor~\cite{thies2016face2face}. Then a 3D model of the target person is built for the face region~\cite{blanz1999morphable}. At test time, given the new expressions, they can be used to drive the 3D model to generate the desired motions, as shown in Figure~\ref{fig:face_reenact_3D}. Instead of extracting the driving expressions from someone else, they can also be directly synthesized from speech inputs~\cite{suwajanakorn2017synthesizing}. Since 3D models are involved, this line of works typically does not use GANs.

Some follow-up works take transferring head motions into account and can model both expressions and different head poses at the same time~\cite{kim2018deep, wu2018reenactgan, bansal2018recycle}. For example, RecycleGAN~\cite{bansal2018recycle} extends CycleGAN~\cite{zhu2017unpaired} to incorporate temporal constraints so it can transform videos of a particular person to another fixed person. On the other hand, ReenactGAN~\cite{wu2018reenactgan} can transfer movements and expressions from an arbitrary person to a fixed person. Still, the subject-dependent nature of these works greatly limits their usability. One model can only work for one person, and generalizing to another person requires training a new model. Moreover, collecting training data for the target person may not be feasible at all times, which motivates the emergence of subject-agnostic models.

\mysubsec{Subject-agnostic.}
Several recent works propose subject-agnostic frameworks, which focus on transferring the facial expressions without head movements~\cite{olszewski2017realistic, chen2018lip, pumarola2018ganimation, nagano2018pagan, geng2018warp, zhou2019talking, chen2019hierarchical, song2019talking, jamaludin2019you, vougioukas2019realistic, fried2019text, pumarola2020ganimation}. In particular, many works only focus on the mouth region, since it is the most expressive part during talking. For example, given an audio speech and one lip image of the target identity, Chen \etal~\cite{chen2018lip} synthesize a video of the desired lip movements. Fried \etal~\cite{fried2019text} edit the lower face region of an existing video, so they can edit the video script and synthesize a new video corresponding to the change. 
While these works have better generalization capability than the previous subject-specific methods, they usually cannot synthesize spontaneous head motions. The head movements cannot be transferred from the driving sequence to the target person.

\begin{figure*}[!t]
    \centering
    \includegraphics[width=.9\textwidth]{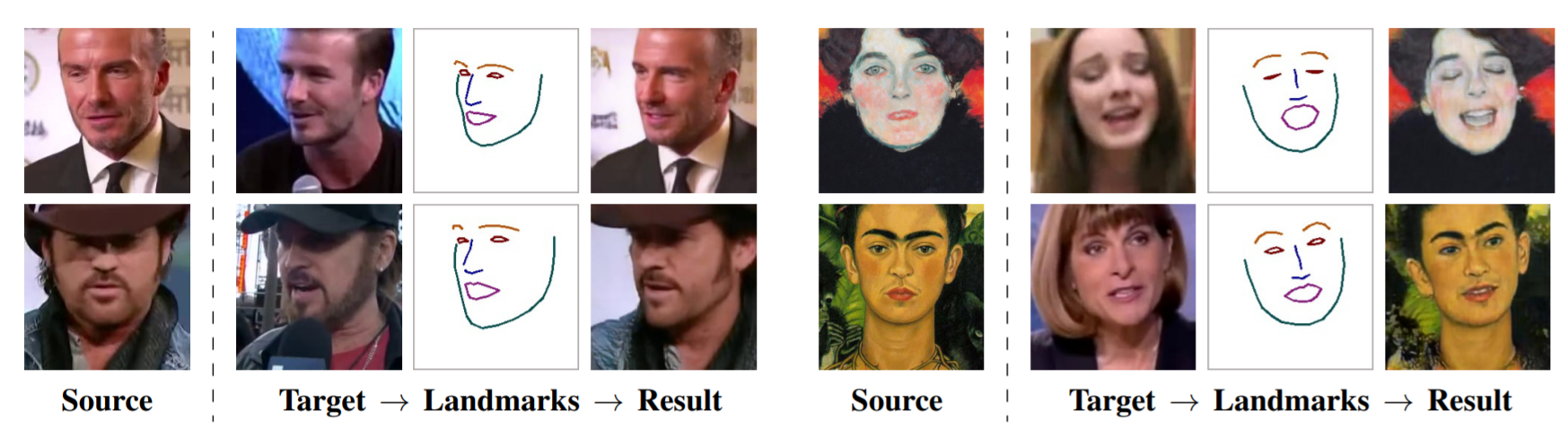}
    \caption{{\bf Few-shot face reenactment methods which require only a 2D image as input~\cite{zakharov2019few}.} The driving expressions are usually represented by facial landmarks or keypoints. Images are from Zakharov \etal~\cite{zakharov2019few}.}
    \label{fig:face_reenact_2D}
\end{figure*}

Some works can very recently handle both expressions and head movements using subject-agnostic frameworks~\cite{averbuch2017bringing, wiles2018x2face, nirkin2019fsgan, zakharov2019few, wang2019few, siarohin2019first, gu2020flnet}. These frameworks only need a single 2D image of the target person and can synthesize talking videos of this person given arbitrary motions. These motions are represented using either facial landmarks~\cite{averbuch2017bringing} or keypoints learned without supervision~\cite{siarohin2019first}. Since the input is only a 2D image, many methods rely on warping the input or its extracted features and then fill in the unoccluded areas to refine the results. For example, Averbuch \etal~\cite{averbuch2017bringing} first warp the image and directly copy the teeth region from the driving image to fill in the holes in case of an open mouth. Siarohin~\etal~\cite{siarohin2019first} warp extracted features from the input image, using motion fields estimated from sparse keypoints. On the other hand, Zakharov~\etal~\cite{zakharov2019few} demonstrate that it is possible to achieve promising results using direct synthesis methods without any warping. To synthesize the target identity, they extract features from the source images and inject the information into the generator through the AdaIN~\cite{huang2017adain} parameters. Similarly, the few-shot vid2vid~\cite{wang2019few} injects the information into their generator by dynamically determining the SPADE~\cite{park2019semantic} parameters. Since these methods require only an image as input, they become particularly powerful and can be used in even more cases. For instance, several works~\cite{averbuch2017bringing, zakharov2019few, wang2019few} demonstrate successes in animating paintings or graffiti instead of real humans, as shown in Figure~\ref{fig:face_reenact_2D}, which is not possible with the previous subject-dependent approaches. However, while these methods have achieved great results in synthesizing people talking under natural motions, they usually struggle to generate satisfying outputs under extreme poses or uncommon expressions, especially when the target pose is very different from the original one.
Moreover, synthesizing complex regions such as hair or background is still hard. This is indeed a very challenging task that is still open to further research. A summary of different categories of face reenactment methods can be found in Table~\ref{tab:face_reenact_category}.

\newcolumntype{M}[1]{>{\centering\arraybackslash}m{#1}}

\begin{table}[!t]
\centering
\caption{{\bf Categorization of face reenactment methods.} Subject-specific models can only work on one subject per model, while subject-agnostic models can work on general targets. Among each of them, some frameworks only focus on the inner face region, so they can only transfer expressions, while others can also transfer head movements. Works with * do not use GANs in their framework.}
\ra{1.3}
\begin{tabular}{M{1cm}M{1.5cm}M{4.5cm}}
\toprule
Target subject            & Transferred region & Methods                    \\ \midrule
\multirow{2}{*}{Specific} & Face only         & \cite{vlasic2005face}*, \cite{thies2015real}*, \cite{thies2016face2face}*, \cite{suwajanakorn2017synthesizing}*, \cite{thies2019deferred}* \\ \cline{2-3} 
                          & Entire head & \cite{kim2018deep}, \cite{wu2018reenactgan, bansal2018recycle}                   \\ \midrule
\multirow{2}{*}{General}  & Face only         & \cite{olszewski2017realistic}, \cite{chen2018lip, pumarola2018ganimation, nagano2018pagan, geng2018warp}, \cite{zhou2019talking}, \cite{chen2019hierarchical, song2019talking}, \cite{jamaludin2019you}*, \cite{vougioukas2019realistic}, \cite{fried2019text}, \cite{pumarola2020ganimation}                   \\ \cline{2-3} 
                          & Entire head & \cite{averbuch2017bringing}*, \cite{wiles2018x2face}*, \cite{nirkin2019fsgan, zakharov2019few}, \cite{qian2019make}*, \cite{wang2019few}, \cite{siarohin2019first}*, \cite{ha2020marionette}          
\\ \bottomrule
\end{tabular}
\label{tab:face_reenact_category}
\end{table}

\subsection{Pose Transfer}
Pose transfer techniques aim at transferring the body pose of one person to another person. It can be seen as the whole body counterpart of face reenactment. In contrast to the talking head generation, which usually shares similar motions, body poses have more varieties and are thus much harder to synthesize. Early works focus on simple pose transfers that generate low resolution and lower quality images. They only work on single images instead of videos. Recent works have shown their capability to generate high quality and high-resolution videos for challenging poses but can only work on a particular person per model. Very recently, several works attempt to perform subject-agnostic video synthesis. A summary of the categories is shown in Table~\ref{tab:pose_transfer_category}. Below we introduce each category in more detail.

\begin{table}[!t]
\centering
\caption{{\bf Categories of pose transfer methods.} Again, they can be classified depending on whether one model can work for only one person or any persons. Some of the frameworks only focus on generating single images, while others also demonstrate their effectiveness on videos. Works with * do not use GANs in their framework.}
\ra{1.3}
\begin{tabular}{M{1cm}M{1.2cm}M{4.8cm}}
\toprule
Target subject           & Output type & Methods  \\ \midrule
Specific                 & Videos       & \cite{thies2018headon}*, \cite{wang2018video}, \cite{chan2019everybody}, \cite{aberman2019deep}, \cite{shysheya2019textured}*, \cite{zhou2019dance}, \cite{liu2019neural}, \cite{liu2020neural} \\ \midrule
\multirow{2}{*}{General} & Images       & \cite{ma2017pose}, \cite{ma2018disentangled}, \cite{siarohin2018deformable}, \cite{esser2018variational}*, \cite{balakrishnan2018synthesizing}, \cite{pumarola2018unsupervised}, \cite{zanfir2018human}*, \cite{joo2018generating, zhao2018multi}, \cite{neverova2018dense, raj2018swapnet}, \cite{dong2018soft}, \cite{song2019unsupervised, grigorev2019coordinate, lorenz2019unsupervised, zhu2019progressive, li2019dense} \\ \cline{2-3} 
                         & Videos       & \cite{yang2018pose}, \cite{siarohin2019monkeynet}, \cite{weng2019photo}*, \cite{liu2019liquid}, \cite{siarohin2019first}*, \cite{wang2019few}, \cite{ren2020deep} \\ \bottomrule
\end{tabular}
\label{tab:pose_transfer_category}
\end{table}

\mysubsec{Subject-agnostic image generation.}
Although we focus on video synthesis in this section, since most of the existing motion transfer approaches only focus on synthesizing images, we still briefly introduce them here (\cite{ma2017pose, ma2018disentangled, siarohin2018deformable, esser2018variational, balakrishnan2018synthesizing, pumarola2018unsupervised, zanfir2018human, joo2018generating, zhao2018multi, neverova2018dense, raj2018swapnet, dong2018soft, song2019unsupervised, grigorev2019coordinate, lorenz2019unsupervised, zhu2019progressive, li2019dense}). Ma \etal~\cite{ma2017pose} adopt a two-stage coarse-to-fine approach using GANs to synthesize a person in a different pose, represented by a set of keypoints. In their follow-up work~\cite{ma2018disentangled}, the foreground, background, and poses in the image are further disentangled into different latent codes to provide more flexibility and controllability. Later, Siarohin \etal~\cite{siarohin2018deformable} introduce deformable skip connections to move local features to the target pose position in a U-Net generator. Similarly, Balakrishnan \etal~\cite{balakrishnan2018synthesizing} decompose different parts of the body into different layer masks and apply spatial transforms to each of them. The transformed segments are then fused together to form the final output.

The above methods work in a supervised setting where images of different poses of the same person are available during training. To work in the unsupervised setting, Pumarola \etal~\cite{pumarola2018unsupervised} render the synthesized image back to the original pose, and apply cycle-consistency constraint on the back-rendered image. Lorenz \etal~\cite{lorenz2019unsupervised} decouple the shape and appearance from images without supervision by adopting a two-stream auto-encoding architecture, so they can re-synthesize images in a different shape with the same appearance.

\begin{figure}[!t]
    \centering
    \includegraphics[width=\columnwidth]{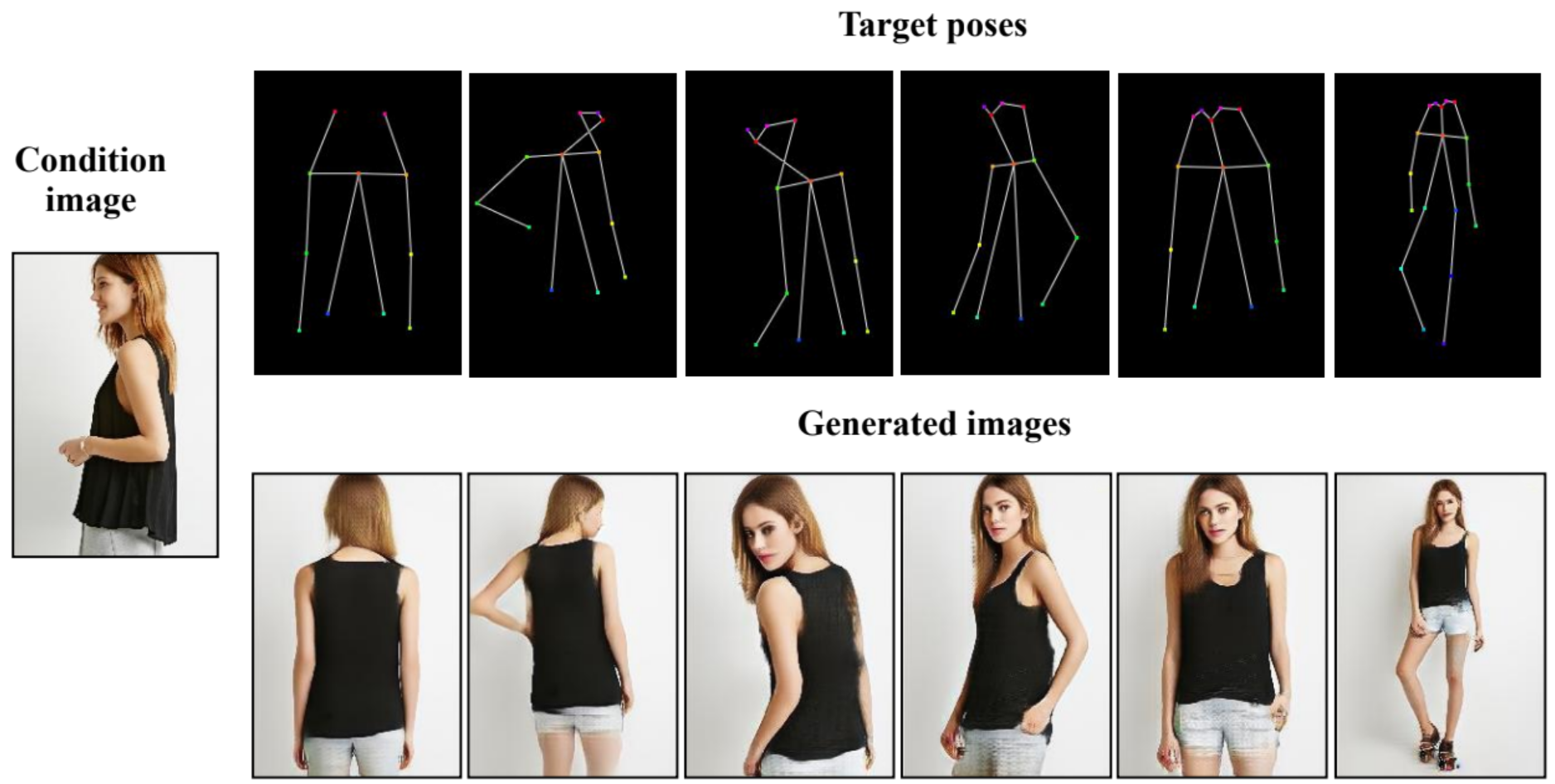}
    \caption{{\bf Subject-agnostic pose transfer examples~\cite{zhu2019progressive}.} Using only a 2D image and the target pose to be synthesized, these methods can realistically generate the desired outputs. Images are from Zhu \etal~\cite{zhu2019progressive}.}
    \label{fig:pose_transfer_general}
\end{figure}

Recently, instead of relying on 2D keypoints solely, some frameworks choose to utilize 3D or 2.5D information. For example, Zanfir \etal~\cite{zanfir2018human} incorporate estimating 3D parametric models into their framework to aid the synthesis process. Similarly, Li \etal~\cite{li2019dense} predict 3D dense flows to warp the source image by estimating 3D models from the input images.
Neverova \etal~\cite{neverova2018dense} adopt the DensePose~\cite{Guler2018DensePose} to help warp the input textures according to their UV-coordinates and inpaint the holes to generate the final result. Grigorev \etal~\cite{grigorev2019coordinate} also map the input to a texture space and inpaint the textures before warping them back to the target pose. Huang \etal~\cite{huang2020arch} combine the SMPL models~\cite{loper2015smpl} with the implicit field estimation framework~\cite{saito2019pifu} to rig the reconstructed meshes with desired motions. While these methods work reasonably well in transferring poses, as shown in Figure~\ref{fig:pose_transfer_general}, directly applying them to videos will usually result in unsatisfactory artifacts such as flickering or inconsistent results. Below we introduce methods specifically targeting video generation, which work on a one-person-per-model basis.

\begin{figure}[!t]
    \centering
    \includegraphics[width=\columnwidth]{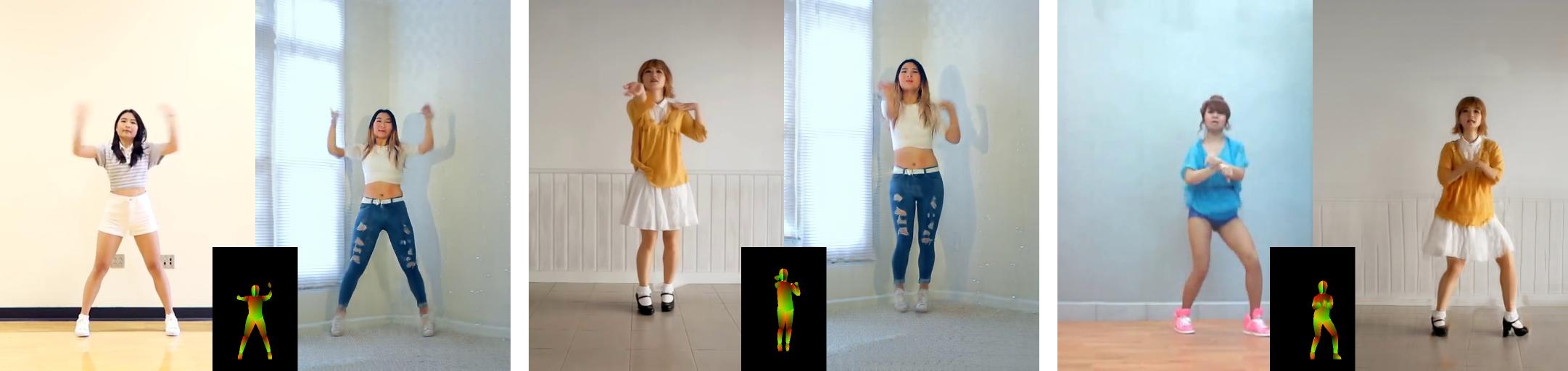}
    \caption{{\bf Subject-specific pose transfer examples for video generation~\cite{wang2018video}}. For each image triplet, left: the driving sequence, middle: the intermediate pose representation, right: the synthesized output. By using a model specifically trained on the target person, it can synthesize realistic output videos faithfully reflecting the driving motions. Images are from Wang \etal~\cite{wang2018video}.}
    \label{fig:pose_transfer_specific}
\end{figure}

\mysubsec{Subject-specific video generation.}
For high-quality video synthesis, most methods employ a subject-specific model, which can only synthesize a particular person. These approaches start with collecting training data of the target person to be synthesized (e.g.\ a few minutes of a subject performing various motions) and then train a neural network or infer a 3D model from it to synthesize the output. For example, Thies \etal~\cite{thies2018headon} extend their previous face reenactment work~\cite{thies2016face2face} to include shoulders and part of the upper body to increase realism and fidelity. To extend to whole-body motion transfer, Wang \etal~\cite{wang2018video} extend their image synthesis framework~\cite{wang2018high} to videos and successfully demonstrate the transfer results on several dancing sequences, opening the era for a new application (Figure~\ref{fig:pose_transfer_specific}). Chan \etal~\cite{chan2019everybody} also adopt a similar approach to generate many dancing examples, but using a simple temporal smoothing on the inputs instead of explicitly modeling temporal consistency by the network. Following these works, many subsequent works improve upon them~\cite{aberman2019deep, shysheya2019textured, zhou2019dance, liu2019neural, liu2020neural}, usually by combining the neural network with 3D models or graphics engines. For example, instead of predicting RGB values directly, Shysheya \etal~\cite{shysheya2019textured} predict DensePose-like part maps and texture maps from input 3D keypoints, and adopt a neural renderer to render the outputs. Liu \etal~\cite{liu2019neural} first construct a 3D character model of the target by capturing multi-view static images and then train a character-to-image translation network using a monocular video of the target. The authors later combine the constructed 3D model with the monocular video to estimate dynamic textures, so they can use different texture maps when synthesizing different motions to increase the realism~\cite{liu2020neural}.

\begin{figure}[!t]
    \centering
    \includegraphics[width=\columnwidth]{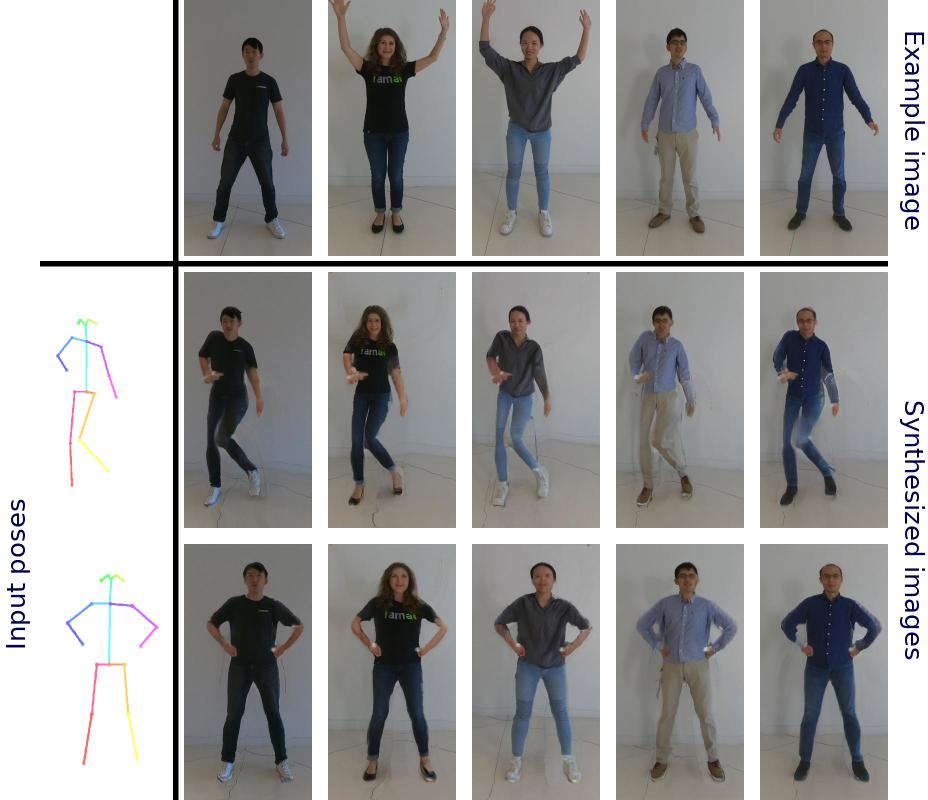}
    \caption{{\bf Subject-agnostic pose transfer videos~\cite{wang2019few}.} Given an example image and a driving pose sequence, the methods can output a sequence of the person performing the motions. Images are from Wang \etal~\cite{wang2019few}.}
    \label{fig:pose_transfer_video}
\end{figure}

\mysubsec{Subject-agnostic video generation.}
Finally, the most general framework would be to have one model that can work universally regardless of the target identity. Early works in this category synthesize videos unconditionally and do not have full control over the synthesized sequence (e.g., MoCoGAN~\cite{tulyakov2018mocogan}). Some other works such as as~\cite{yang2018pose} have control over the appearance and the starting pose of the person, but the motion generation is still unconditional. Due to these factors, the synthesized videos are usually shorter and of lower quality. Very recently, a few works have shown the ability to render higher quality videos for pose transfer results~\cite{weng2019photo, liu2019liquid, wang2019few, siarohin2019monkeynet, siarohin2019first, ren2020deep}. Weng \etal~\cite{weng2019photo} reconstruct the SMPL model~\cite{loper2015smpl} from the input image and animate it with some simple motions like running. Liu \etal~\cite{liu2019liquid} propose a unified framework for pose transfer, novel view synthesis, and appearance transfer all at once. Siarohin  \etal~\cite{siarohin2019monkeynet, siarohin2019first} estimate unsupervised keypoints from the input images and predict a dense motion field to warp the source features to the target pose. Want \etal~\cite{wang2019few} extend vid2vid~\cite{wang2018video} to the few-shot setting by predicting kernels in the SPADE~\cite{park2019semantic} modules. Similarly, Ren \etal~\cite{ren2020deep} also predict kernels in their local attention modules using the input images to adaptively select features and warp them to the target pose. While these approaches have achieved better results than previous works (Figure~\ref{fig:pose_transfer_video}), their qualities are still not comparable to state-of-the-art subject-specific models. Moreover, most of them still synthesize lower resolution outputs ($256$ or $512$). How to further increase the quality and resolution to the photorealistic level is still an open question.

\section{Neural Rendering}
\label{sec:neural_rendering}


\begin{figure*}
  \centering
  \includegraphics[width=.95\linewidth]{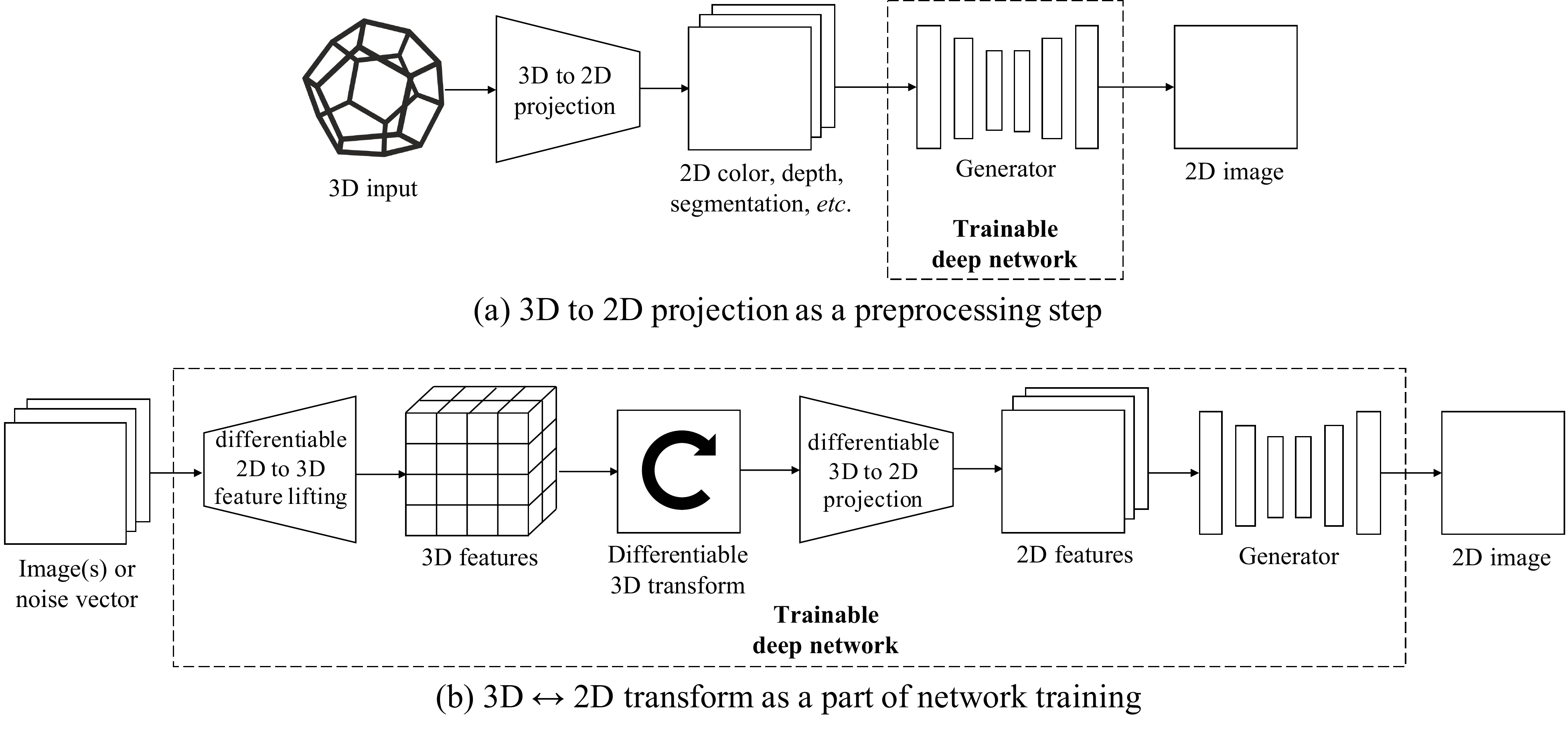}
  \vspace{-2mm}
  \caption{\textbf{The two common frameworks for neural rendering.} (a) In the first set of works~\cite{brualla2018lookinggood,mallya2020world,meshry2019neural,pittaluga2019revealing,li2020crowdsampling}, a neural network that purely operates in the 2D domain is trained to enhance an input image, possibly supplemented with other information such as depth, or segmentation maps. 
  (b) The second set of works~\cite{sitzmann2019deepvoxels,nguyen2019hologan,nguyen2020blockgan,wiles2019synsin,schwarz2020graf} introduces native 3D operations that produce and transform 3D features. This allows the network to reason in 3D and produce view-consistent outputs.}
\label{figs:neural_rendering:frameworks}
\end{figure*}

Neural rendering is a recent and upcoming topic in the area of neural networks, which combines classical rendering and generative models. Classical rendering can produce photorealistic images given the complete specification of the world. This includes all the objects in it, their geometry, material properties, the lighting, the cameras, etc. Creating such a world from scratch is a laborious process that often requires expert manual input. Moreover, faithfully reproducing such data directly from images of the world can often be hard or impossible. On the other hand, as described in the previous sections, GANs have had great success in producing photorealistic images given minimal semantic inputs. The ability to synthesize and learn material properties, textures, and other intangibles from training data can help overcome the drawbacks of classical rendering.

Neural rendering aims to combine the strengths of the two areas to create a more powerful and flexible framework. Neural networks can either be applied as a postprocessing step after classical rendering or as part of the rendering pipeline with the design of 3D-aware and differentiable layers. The following sections discuss such approaches and how they use GAN losses to improve the quality of outputs. In this paper, we focus on works that use GANs to train neural networks and augment the classical rendering pipeline to generate images. For a general survey on the use of neural networks in rendering, please refer to the survey paper on neural rendering by Tewari \etal~\cite{tewari2020neuralSTAR}.

We divide the works on GAN-based neural rendering into two parts: 1) works that treat 3D to 2D projection as a preprocessing step and apply neural networks purely in the 2D domain, and 2) works that incorporate layers that perform differentiable operations to transform features from 3D to 2D or vice versa (3D $\leftrightarrow$ 2D) and learn some implicit form of geometry to provide 3D understanding to the network.

\subsection{\textbf{3D to 2D projection as a preprocessing step}}
\label{subsec:3dto2dpreprocessing}

A number of works~\cite{brualla2018lookinggood,mallya2020world,meshry2019neural,pittaluga2019revealing,li2020crowdsampling} improve upon traditional techniques by casting the task of rendering into the framework of image-to-image translation, possibly unimodal, multimodal, or conditional, depending on the exact use-case. Using given camera parameters, the source 3D world is first projected to a 2D feature map containing per-pixel information such as color, depth, surface normals, segmentation, \etc. This feature map is then fed as input to a generator, which tries to produce desired outputs, usually a realistic-looking RGB image. The deep neural network application happens in the 2D space after the 3D world is projected to the camera view, and no features or gradients are backpropagated to the 3D source world or through the camera projection. A key advantage of this approach is that the traditional graphics rendering pipeline can be easily augmented to immediately take advantage of proven and mature techniques from 2D image-to-image translation (as discussed in Section~\ref{sec:supervised}), without the need for designing and implementing differentiable projection layers or transformations that are part of the deep network during training. This type of framework is illustrated in Figure~\ref{figs:neural_rendering:frameworks} (a).

Martin-Brualla \etal~\cite{brualla2018lookinggood} introduce the notion of re-rendering, where a deep neural network takes as input a rendered 2D image and enhances it (improving colors, boundaries, resolution, \etc) to produce a re-rendered image. The full pipeline consists of two steps---a traditional 3D to 2D rendering step and a trainable deep network that enhances the rendered 2D image. The 3D to 2D rendering technique can be differentiable or non-differentiable, but no gradients are backpropagated through this step. This allows one to use more complex rendering techniques.
By using this two-step process, the output of a performance capture system, which might suffer from noise, poor color reproduction, and other issues, can be improved. In this particular work, they did not see an improvement from using a GAN loss, perhaps because they trained their system on the limited domain of people and faces, using carefully captured footage.

Meshry \etal~\cite{meshry2019neural} and Li \etal~\cite{li2020crowdsampling} extend this approach to the more challenging domain of unstructured photo collections. They produce multiple plausible views of famous landmarks from noisy point clouds generated from internet photo collections by utilizing Structure from Motion (SfM).  Meshry \etal~\cite{meshry2019neural} generate a 2D feature map containing per-pixel albedo and depth by splatting points of the 3D point cloud onto a given viewpoint. The segmentation map of the expected output image is also concatenated to this feature representation. The problem is then framed as a multimodal image translation problem. A noisy and incomplete input has to be translated to a realistic image conditioned on a style code to produce desired environmental effects such as lighting. 
Li \etal~\cite{li2020crowdsampling} use a similar approach, but with multi-plane images and achieve better photo-realism.
Pittaluga \etal~\cite{pittaluga2019revealing} tackle the task of producing 2D color images of the underlying scene given as input a sparse SfM point cloud with associated point attributes such as color, depth, and SIFT descriptors. 
The input to their network is a 2D feature map obtained by projecting the 3D points to the image plane given the camera parameters. The attributes of the 3D point are copied to the 2D pixel location to which it is projected. Mallya \etal~\cite{mallya2020world} precompute the mapping of the 3D world point cloud to the pixel locations in the images produced by cameras with known parameters and use this to obtain an estimate of the next frame, referred to as a `\emph{guidance image}'. They learn to output video frames consistent over time and viewpoints by conditioning the generator on these noisy estimates.

\begin{figure}
  \centering
  \includegraphics[width=0.8\linewidth]{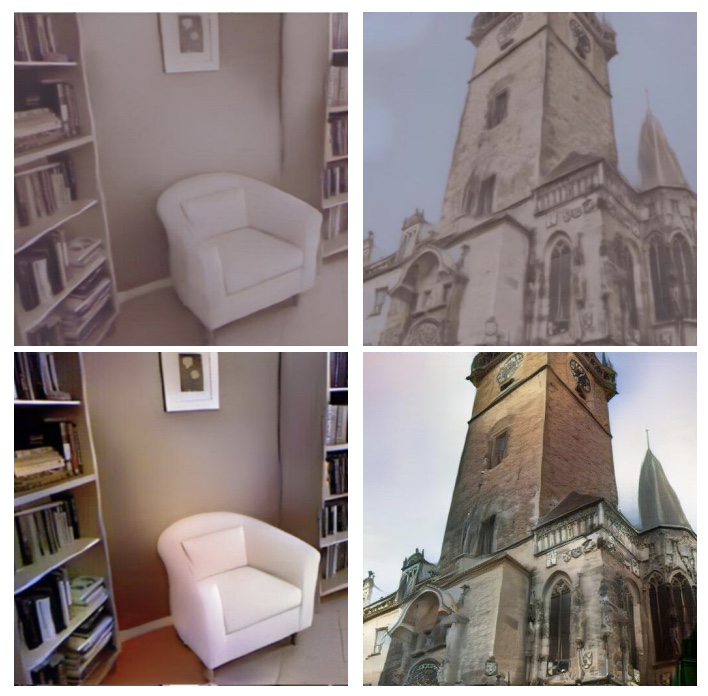}
  \caption{\textbf{Inverting images from 3D point clouds and their associated depth and SIFT attributes~\cite{pittaluga2019revealing}.} The top row of images are produced by a generator trained without an adversarial loss, while the bottom row uses adversarial loss. Using an adversarial loss helps generates better details and more plausible colors.
  Images are from Pittaluga \etal~\cite{pittaluga2019revealing}.}
\label{figs:neural_rendering:pittaluga2019revealing:outputs}
\end{figure}

In these works, the use of a generator coupled with an adversarial loss helps produce better-looking outputs conditioned on the input feature maps. Similar to applications of pix2pixHD~\cite{wang2018high}, such as manipulating output images by editing input segmentation maps, Meshry \etal~\cite{meshry2019neural} are able to remove people and transient objects from images of landmarks and generate plausible inpainting. A key motivation of the work of Pittaluga \etal~\cite{pittaluga2019revealing} was to explore if a user's privacy can be protected by techniques such as discarding the color of the 3D points. A very interesting observation was that discarding color information helps prevent accurate reproduction. However, the use of a GAN loss recovers plausible colors and greatly improves the output results, as shown in Figure.~\ref{figs:neural_rendering:pittaluga2019revealing:outputs}. GAN losses might also be helpful in cases where it is hard to manually define a good loss function, either due to the inherent ambiguity in determining the desired behavior or the difficulty in fully labeling the data.


\subsection{\textbf{3D $\leftrightarrow$ 2D transform as a part of network training}}
\label{subsec:3dto2dtransform}

In the previous set of works, the geometry of the world or object is explicitly provided, and neural rendering is purely used to enhance the appearance or add details to the traditionally rendered image or feature maps. The works in this section~\cite{sitzmann2019deepvoxels,nguyen2019hologan,nguyen2020blockgan,wiles2019synsin,schwarz2020graf} introduce native 3D operations in the neural network used to learn from and produce images. 
These operations enable them to model the geometry and appearance of the scene in the feature space. The general pipeline of this line of works is illustrated in Figure~\ref{figs:neural_rendering:frameworks} (b).
Learning a 3D representation and modeling the process of image projection and formation into the network have several advantages: the ability to reason in 3D, control the pose, and produce a series of consistent views of a scene. Contrast this to the neural network shown in Figure~\ref{figs:neural_rendering:frameworks} (a), which purely operates in the 2D domain.

DeepVoxels~\cite{sitzmann2019deepvoxels} learns a persistent 3D voxel feature representation of a scene given a set of multi-view images and their associated camera intrinsic and extrinsic parameters. Features are first extracted from the 2D views and then lifted to a 3D volume. This 3D volume is then integrated into the persistent DeepVoxels representation. These 3D features are then projected to 2D using a projection layer, and a new view of the object is synthesized using a U-Net generator. This generator network is trained with an \(\ell_1\) loss and a GAN loss. The authors found that using a GAN loss accelerates the generation of high-frequency details, especially at earlier stages of training. Similar to DeepVoxels~\cite{sitzmann2019deepvoxels}, Visual Object Networks (VONs)~\cite{zhu2018visual} generate a voxel grid from a sample noise vector and use a differentiable projection layer to map the voxel grid to a 2.5D sketch. Inspired by classical graphics rendering pipelines, this work decomposes image formation into three conditionally independent factors of shape, viewpoint, and texture. Trained with a GAN loss, their model synthesizes more photorealistic images, and the use of the disentangled representation allows for 3D manipulations, which are not feasible with purely 2D methods.

HoloGAN~\cite{nguyen2019hologan} proposes a system to learn 3D voxel feature representations of the world and to render it to realistic-looking images. Unlike VONs~\cite{zhu2018visual}, HoloGAN does not require explicit 3D data or supervision and can do so using unlabeled images (no pose, explicit 3D shape, or multiple views). By incorporating a 3D rigid-body transformation module and a 3D-to-2D projection module in the network,  HoloGAN provides the ability to control the pose of the generated objects. HoloGAN employs a multi-scale feature GAN discriminator, and the authors empirically observed that this helps prevent mode collapse. BlockGAN~\cite{nguyen2020blockgan} extends the unsupervised approach of the HoloGAN~\cite{nguyen2019hologan} to also consider object disentanglement. BlockGAN learns 3D features per object and the background. These are combined into 3D scene features after applying appropriate transformations before projecting them into the 2D space. One issue with learning scene compositionality without explicit supervision is the conflation of features of the foreground object and the background, which results in visual artifacts when objects or the camera moves. By adding more powerful `style' discriminators (feature discriminators introduced in~\cite{nguyen2019hologan}) to their training scheme, the authors observed that the disentangling of features improved, resulting in cleaner outputs.

SynSin~\cite{wiles2019synsin} learns an end-to-end model for view
synthesis from a single image, without any ground-truth 3D supervision. Unlike the above works which internally use a feature voxel representation, SynSin predicts a point cloud of features from the input image and then projects it to new views using a differentiable point cloud renderer. 2D image features and a depth map are first predicted from the input image. Based on the depth map, the 2D features are projected to 3D to obtain the 3D feature point cloud. The network is trained adversarially with a discriminator based on the one proposed by Wang \etal~\cite{wang2018high}.

One of the drawbacks of voxel-based feature representations is the cubic growth in the memory required to store them. To keep requirements manageable, voxel-based approaches are typically restricted to low resolutions. GRAF~\cite{schwarz2020graf} proposes to use conditional radiance fields, which are a continuous mapping from a 3D location and a 2D viewing direction to an RGB color value, as the intermediate feature representation. They also use a single discriminator similar to PatchGAN~\cite{isola2017image}, with weights that are shared across patches with different receptive fields. This allows them to capture the global context as well as refine local details.


\begin{table}[t]
    \ra{1.3}
	\centering
	\caption{\textbf{Key differences amongst 3D-aware methods}. Adversarial losses are used by a range of methods that differ in the type of 3D feature representation and training supervision.}
	\begin{tabular}{M{2cm}M{2cm}M{2.5cm}}
		\toprule
		3D feature representation & Supervision & Methods \\ \midrule
		Radiance field & \multirow{3}{*}{None} & GRAF~\cite{schwarz2020graf} \\ \cline{1-1} \cline{3-3}
		
		\multirow{5}{*}{Voxel} & & HoloGAN~\cite{nguyen2019hologan}\\
		& & BlockGAN~\cite{nguyen2020blockgan} \\ \cline{2-3}

		& 3D supervision & VONs~\cite{zhu2018visual} \\ \cline{2-3}

		& Input-Output & DeepVoxels~\cite{sitzmann2019deepvoxels}\\ \cline{1-1} \cline{3-3}

		Point cloud & pose transformation & SynSin~\cite{wiles2019synsin} \\ 
		\bottomrule
	\end{tabular}
	\label{tab:neural_rendering:comparison}
\end{table}

As summarized in Table~\ref{tab:neural_rendering:comparison}, the works discussed in this section use a variety of 3D feature representations, and train their networks using paired input-output with known transformations or unlabeled and unpaired data. The use of a GAN loss is common to all these approaches. This is perhaps because traditional hand-designed losses such as the \(\ell_1\) loss or even perceptual loss are unable fully to capture what makes a synthesized image look unrealistic. Further, in the case where explicit task supervision is unavailable, BlockGAN~\cite{nguyen2020blockgan} shows that a GAN loss can help in learning disentangled features by ensuring that the outputs after projection and rendering look realistic. The learnability and flexibility of the GAN loss to the task at hand helps provide feedback, guiding how to change the generated image, and thus the upstream features, so that it looks as if it were sampled from the distribution of real images. This makes the GAN framework a powerful asset in the toolbox of any neural rendering practitioner.
\section{Limitations and Open Problems}\label{sec:limitations}
Despite the successful applications introduced above, there are still limitations of GANs needed to be addressed by future work.

\mysubsec{Evaluation metrics.} Evaluate and comparing different GAN models is difficult. The most popular evaluation metrics are perhaps Inception Score~(IS)~\cite{salimans2016improved} and Fréchet Inception Distance~(FID)~\cite{heusel2017gans}, which both have many shortcomings. The Inception Score, for example, is not able to detect intra-class mode collapse~\cite{borji2019pros}. In other words, a model that generates only a single image per class can obtain a high IS. FID can better measure such diversity, but it does not have an unbiased estimator~\cite{binkowski2018demystifying}. Kernel Inception Distance~(KID)~\cite{binkowski2018demystifying} can capture higher-order statistics and has an unbiased estimator but has been empirically found to suffer from high variance~\cite{ravuri2019seeing}. In addition to the above measures that summarize the performance with a single number, there are metrics that separately evaluate fidelity and diversity of the generator distribution~\cite{sajjadi2018assessing,kynkaanniemi2019improved,ferjad2020icml}.

\mysubsec{Instability.} Although the regularization techniques introduced in section \ref{sec:reg} have greatly improved the stability of GAN training, GANs are still much more unstable to train than supervised discriminative models or likelihood-based generative models. For example, even the state-of-the-art BigGAN model would eventually collapse in the late stage of training on ImageNet~\cite{brock2018large}. Also, the final performance is generally very sensitive to hyper-parameters~\cite{lucic2018gans,kurach2019large}.

\mysubsec{Interpretability.} Despite the impressive quality of the generate images, there has been a lack of understanding of how GANs represent the image structure internally in the generator. Bau~\etal visualize the causal effect of different neurons on the output image~\cite{bau2019gan}. After finding the semantic meaning of individual neurons or directions in the latent space~\cite{jahanian2020steerability,goetschalckx2019ganalyze,shen2020interpreting}, one can edit a real image by inverting it to the latent space, edit the latent code according to the desired semantic change, and regenerate it with the generator. Finding the best way to encode an image to the latent space is, therefore, another interesting research direction~\cite{zhu2016generative,abdal2019image2stylegan,abdal2020image2stylegan2,bau2019seeing,karras2020analyzing,huh2020transforming}. 

\mysubsec{Forensics.} The success of GANs has enabled many new applications but also raised ethical and social concerns such as fraud and fake news. The ability to detect GAN-generated images is essential to prevent malicious usage of GANs. Recent studies have found it possible to train a classifier to detect generated images and generalize to unseen generator architectures~\cite{zhang2019detecting,wang2020cnn,chai2020makes}. This cat-and-mouse game may continue, as generated images may become increasingly harder to detect in the future. 
\section{Conclusion}\label{sec:conclusion}
In this paper, we present a comprehensive overview of GANs with an emphasis on algorithms and applications to visual synthesis. We summarize the evolution of the network architectures in GANs and the strategies to stabilize GAN training. We then introduce several fascinating applications of GANs, including image translation, image processing, video synthesis, and neural rendering.
In the end, we point out some open problems for GANs, and we hope this paper would inspire future research to solve them.



\ifCLASSOPTIONcaptionsoff
  \newpage
\fi



%
%
%

{\small
	\bibliographystyle{ieee}
    \bibliography{bib/combined}
}

%
\begin{IEEEbiography}[{\includegraphics[width=1in,height=1.25in,clip,keepaspectratio]{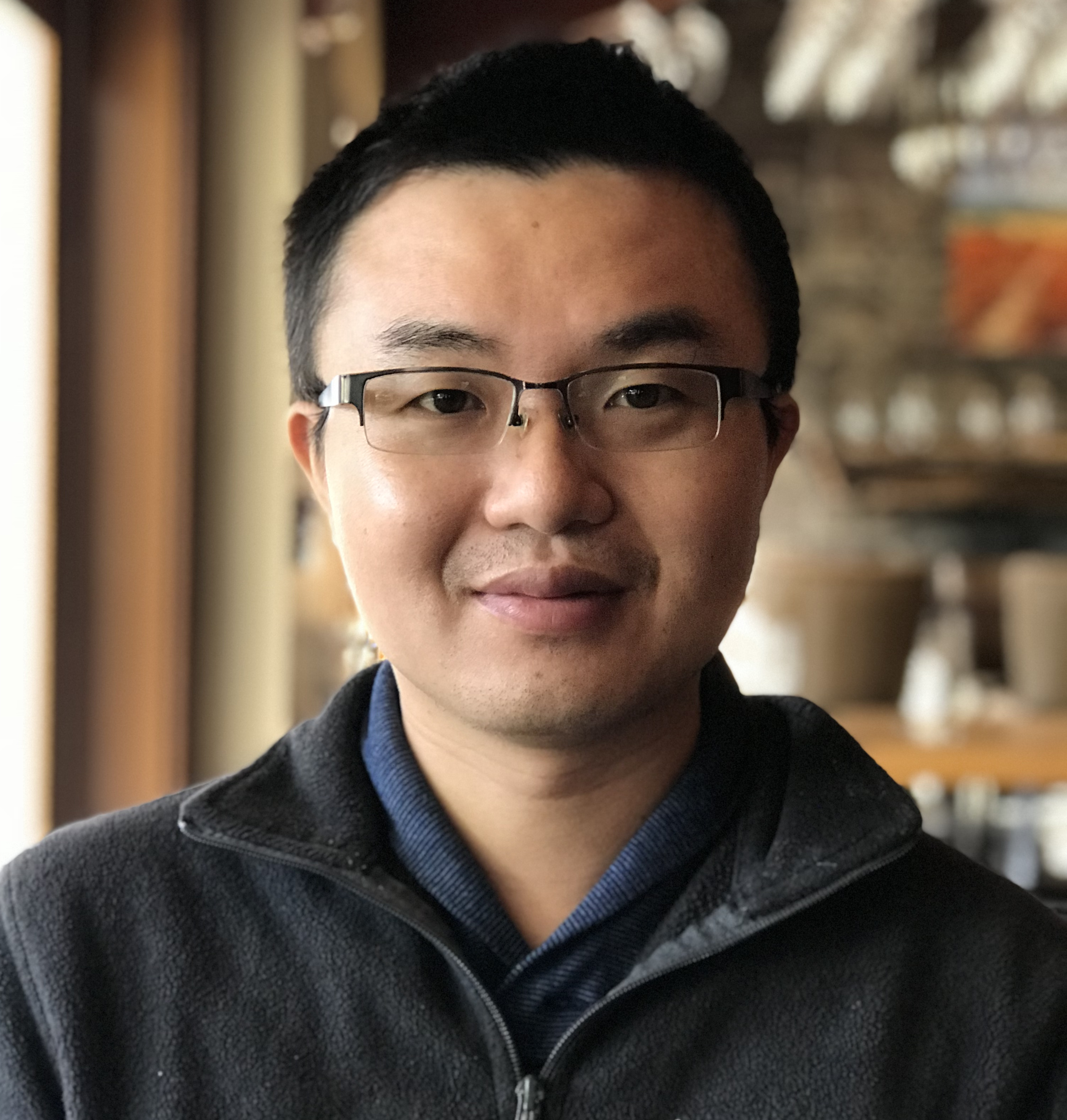}}]{Ming-Yu Liu}
is a Distinguished Research Scientist and Manager at NVIDIA Research. Before joining NVIDIA in 2016, he was a Principal Research Scientist at Mitsubishi Electric Research Labs (MERL). He received his Ph.D. from the Department of Electrical and Computer Engineering at the University of Maryland College Park in 2012. Ming-Yu Liu has won several prestigious awards in his field. He is a recipient of the R\&D 100 Award by R\&D Magazine in 2014 for his robotic bin picking system. In SIGGRAPH 2019, he won the Best in Show Award and Audience Choice Award in the Real-Time Live track for his GauGAN work. His GauGAN work also won the Best of What's New Award by the Popular Science Magazine in 2019. His research interest is on generative image modeling. His goal is to enable machines human-like imagination capability.

\end{IEEEbiography}
\begin{IEEEbiography}[{\includegraphics[width=1in,height=1.25in,clip,keepaspectratio]{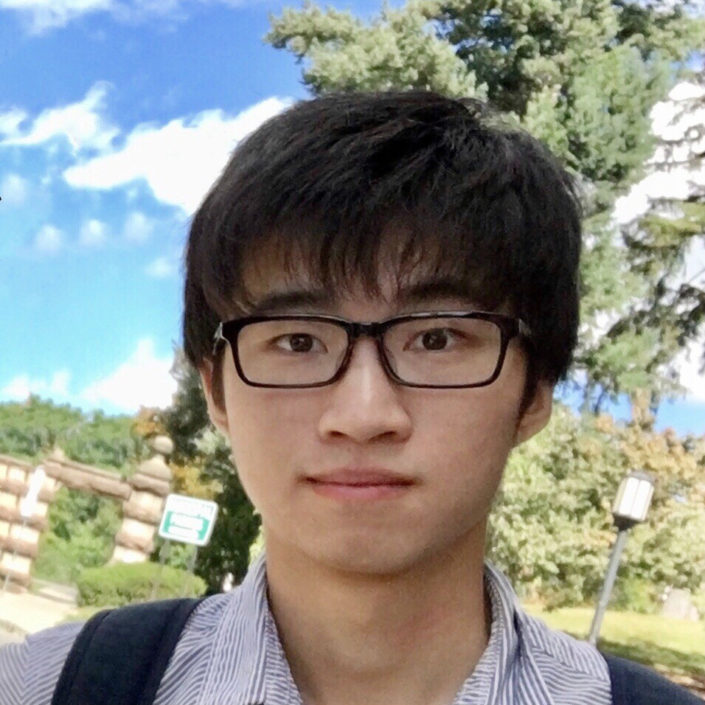}}]{Xun Huang}
is a Research Scientist at NVIDIA Research. He obtained his Ph.D. from Cornell University under the supervision of Professor Serge Belongie. His research interest includes developing new architectures and training algorithms of generative adversarial networks, as well as applications such as image editing and synthesis. He is a recipient of NVIDIA Graduate Fellowship, Adobe Research Fellowship, and Snap Research Fellowship.
\end{IEEEbiography}

\begin{IEEEbiography}[{\includegraphics[width=1in,height=1.25in,clip,keepaspectratio]{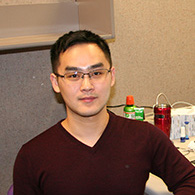}}]{Jiahui Yu}
is a Research Scientist at Google Brain. He received his Ph.D. at University of Illinois at Urbana-Champaign in 2020, and Bachelor with distinction at School of the Gifted Young in Computer Science, University of Science and Technology of China in 2016. His research interest is in sequence modeling (language, speech, video, financial data), machine perception (vision), generative models (GANs), and high performance computing. He is a member of IEEE, ACM and AAAI. He is a recipient of Baidu Scholarship, Thomas and Margaret Huang Research Award, and Microsoft-IEEE Young Fellowship.
\end{IEEEbiography}

\begin{IEEEbiography}[{\includegraphics[width=1in,height=1.25in,clip,keepaspectratio]{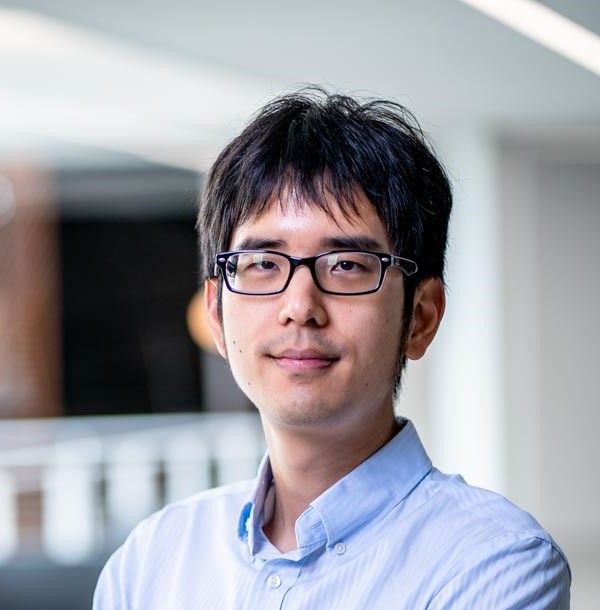}}]{Ting-Chun Wang}
is a senior research scientist at NVIDIA Research. He obtained his Ph.D. in EECS from UC Berkeley, advised by Professor Ravi Ramamoorthi and Alexei A. Efros. He won the 1st place in the Domain Adaptation for Semantic Segmentation Competition in CVPR, 2018. His semantic image synthesis paper was in the best paper finalist in CVPR, 2019, and the corresponding GauGAN app won the Best in Show Award and Audience Choice Award in SIGGRAPH RealTimeLive, 2019. He served as an area chair in WACV, 2020. His research interests include computer vision, machine learning and computer graphics, particularly the intersections of all three. His recent research focus is on using generative adversarial models to synthesize realistic images and videos, with applications to rendering, visual manipulations and beyond.
\end{IEEEbiography}

\begin{IEEEbiography}[{\includegraphics[width=1in,height=1.25in,clip,keepaspectratio]{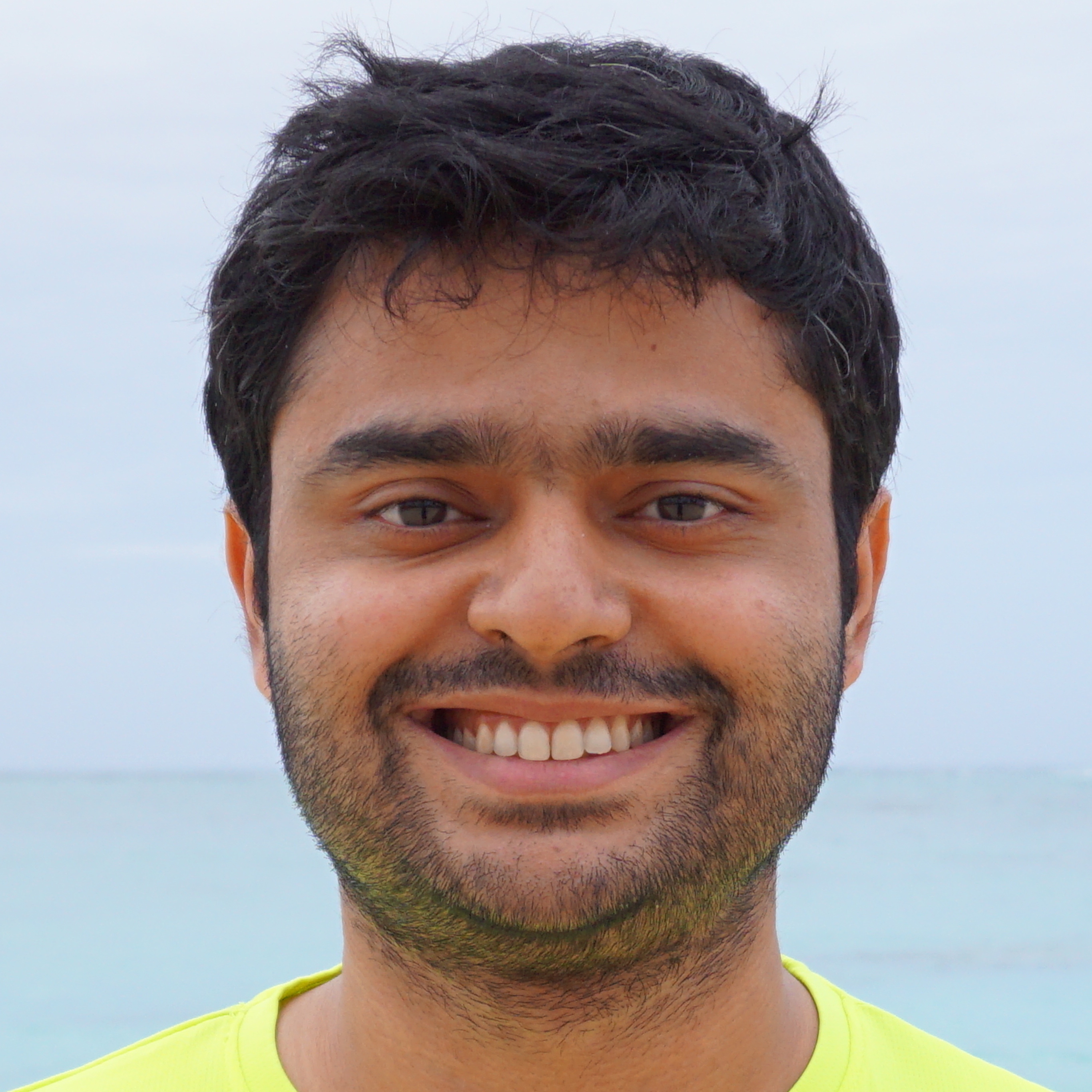}}]{Arun Mallya}
is a Senior Research Scientist at NVIDIA Research. He obtained his Ph.D. from the University of Illinois at Urbana-Champaign in 2018, with a focus on performing multiple tasks efficiently with a single deep network. He holds a B.Tech. in Computer Science and Engineering from the Indian Institute of Technology - Kharagpur (2012), and an MS in Computer Science from the University of Illinois at Urbana-Champaign (2014). He was selected as a Siebel Scholar in 2014. He is interested in generative modeling and enabling new applications of deep neural networks.
\end{IEEEbiography}







\end{document}